\documentclass[10pt,twocolumn,letterpaper]{article}

\usepackage[pagenumbers]{cvpr} 
\usepackage{booktabs}
\usepackage{array}
\usepackage{pifont}
\usepackage{multirow}
\usepackage{graphicx}
\usepackage{xcolor,colortbl}
\usepackage{algorithm}
\usepackage{algpseudocode}

\newcolumntype{Y}{>{\raggedright\arraybackslash}X}
\definecolor{cvprblue}{rgb}{0.21,0.49,0.74}
\usepackage[pagebackref,breaklinks,colorlinks,allcolors=cvprblue]{hyperref}

\def\paperID{} 
\def\confName{CVPR}
\def\confYear{2026}

\title{SkyLume: A Large-Scale Multi-Illumination Aerial Benchmark for Urban Scene Reconstruction and Beyond}

\author{
Zhuoxiao Li$^1$\quad Wenzong Ma$^1$\quad Taoyu Wu$^2$\quad Jinjing Zhu$^1$\quad Shuai Zhang$^1$\quad Jing OU$^1$\quad \\ \quad Tongyan Hua$^1$\quad Yinrui Ren$^1$ \quad Rongjun Qin$^3$ \quad  Hui Xiong$^1$\quad Wufan Zhao$^1$
 \\
{\normalsize $^1$HKUST(GZ), $^2$University of Liverpool, $^3$The Ohio State University} \\
}

\begin{document}
\twocolumn[{
    \maketitle
    \begin{center}
        \vspace{-20pt}
        \includegraphics[width=1\textwidth]{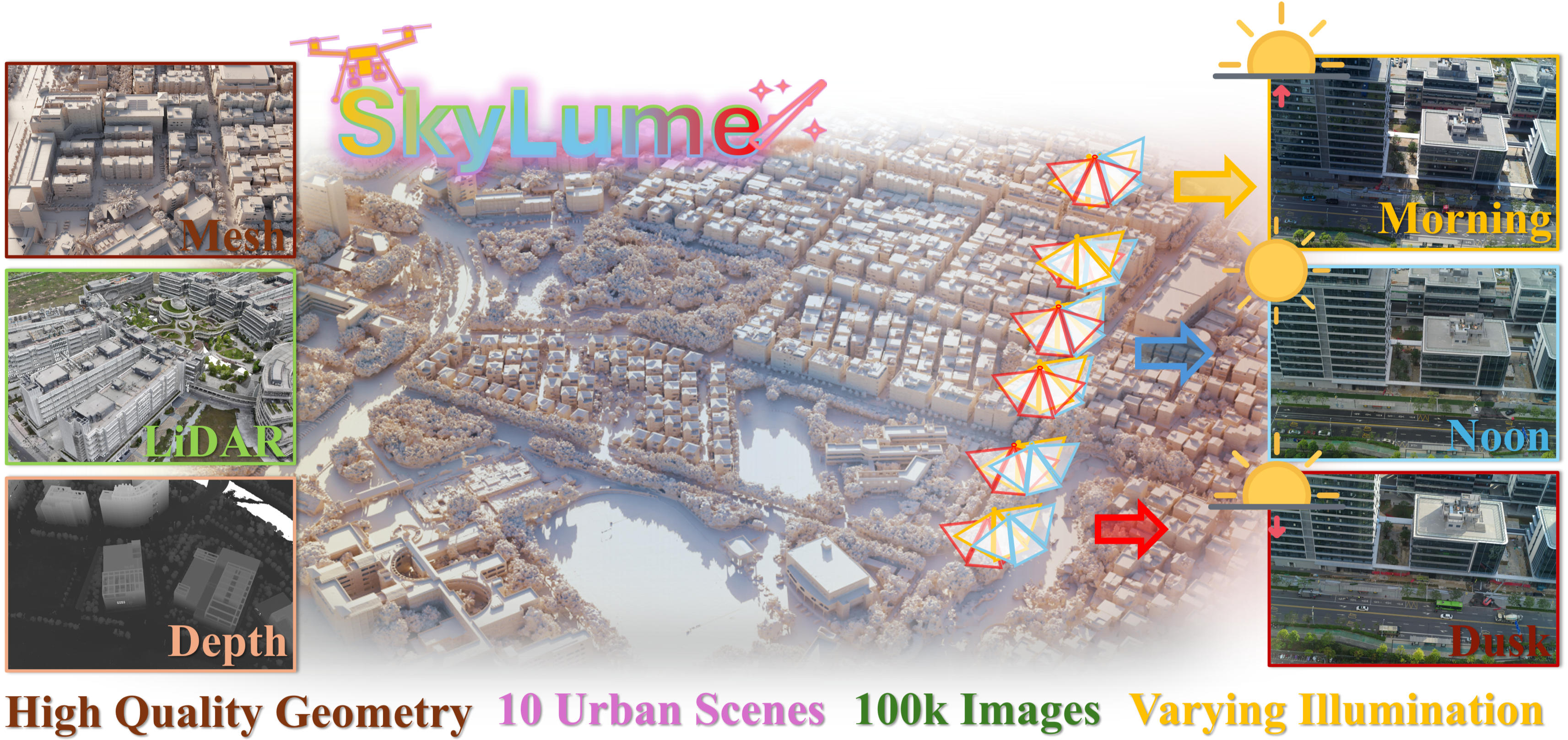}
        \vspace{-15pt}
        \captionof{figure}{We present \textbf{SkyLume}, the first comprehensive real-world UAV dataset centered on illumination variation. It provides 6K-resolution five-direction imagery from three daily captures along identical RTK-guided flight paths, paired with LiDAR-derived ground truth including precise meshes and per-frame depth and normals under unified 6-DoF poses. The benchmark enables rigorous evaluation of 3D reconstruction, novel view synthesis. Moreover, leveraging the three time-of-day captures, \textbf{SkyLume} is the first real-world UAV dataset to enable rigorous city-scale evaluation of inverse-rendering quality. }
        \label{fig:teaser}
    \end{center}
}]
\maketitle

\begin{abstract}

Recent advances in Neural Radiance Fields and 3D Gaussian Splatting have demonstrated strong potential for large-scale UAV-based 3D reconstruction tasks by fitting the appearance of images. However, real-world large-scale captures are often based on multi-temporal data capture, where illumination inconsistencies across different times of day can significantly lead to color artifacts, geometric inaccuracies, and inconsistent appearance. Due to the lack of UAV datasets that systematically capture the same areas under varying illumination conditions, this challenge remains largely underexplored. To fill this gap, we introduce \textbf{SkyLume}, a large-scale, real-world UAV dataset specifically designed for studying illumination robust 3D reconstruction in urban scene modeling:
(1) We collect data from 10 urban regions data comprising more than 100k high resolution UAV images (four oblique views and nadir), where each region is captured at three periods of the day to systematically isolate illumination changes. (2) To support precise evaluation of geometry and appearance, we provide per-scene LiDAR scans and accurate 3D ground-truth for assessing depth, surface normals, and reconstruction quality under varying illumination. (3) For the inverse rendering task, we introduce the Temporal Consistency Coefficient (TCC), a metric that measuress cross-time albedo stability and directly evaluates the robustness of the disentanglement of light and material. We aim for this resource to serve as a foundation that advances research and real-world evaluation in large-scale inverse rendering, geometry reconstruction, and novel view synthesis. Project Page: \href{https://zhuoxiaoli.github.io/skylume_page}{skylume}.
\end{abstract}    
\section{Introduction}
\label{sec:intro}

UAV oblique photogrammetry has rapidly become a practical tool for city-scale modeling and visualization~\cite{zhou202uavphoto}. As 3D vision technologies mature, structure-from-motion (SfM)~\cite{schonberger2016sfm} and multi-view stereo (MVS)~\cite{seitz2006mvs} have scaled through improved matching and learned regularization, while neural radiance fields (NeRF)~\cite{mildenhall2021nerf} and 3D Gaussian Splatting (3DGS)~\cite{kerbl20233d} advance continuous-view rendering, achieving higher fidelity and faster training. These advancements improve geometric completeness, lighting realism, and interactive performance, bringing urban digital twins closer to reality~\cite{meganerf, chen2024gigags, vastgaussian, mi2023switchnerf, chen2024dogaussiandistributedorientedgaussiansplatting, liu2024citygaussianv2efficientgeometricallyaccurate}. However, in UAV-based 3D vision, particularly at the city scale, significant challenges arise due to multi-temporal data capture. Flights are typically conducted at different times of day under varying lighting conditions, which complicates the construction of globally consistent models. Current pipelines often integrate observed lighting into textures or radiance fields but fail to maintain a consistent appearance over time~\cite{zhang2025ref-gs, yao2024ref-gaussian, albedo_isprs}. This raises a central question: \textbf{\textit{Can current methods preserve geometric fidelity, photometric consistency, and novel-view realism under illumination shifts across different flight sessions?}}

To address this question, it is essential to assess the impact of varying illumination conditions on the same scene, as this evaluation is crucial for advancing and refining existing methodologies. Illumination is a critical factor beyond aesthetics~\cite{albedo_isprs, xie2025envgs}, as shadow and exposure variations alter keypoint statistics and multi-view correspondences, which can bias depth and normal estimates~\cite{langguth2016shading, philip2019shadow}. However, existing UAV datasets do not sufficiently address this need. Many datasets provide either synthetic illumination diversity based on simplified light transport models from game engines~\cite{li2023matrixcity, zhao2024aerialgo}, or real-world data that seldom revisits the same regions at different times of the day~\cite{xiong2024gauu, lin2022us3d, jiang2025horizon, wang2025uavscenes}. The lack of consistent illumination changes across different time slots and geographic regions limits the ability to test the stability of existing methods under real-world dynamic lighting conditions. Furthermore, the existing datasets either fail to include adequate ground-truth geometry~\cite{meganerf, isprs-annals-II-3-W4-135-2015, jiang2025horizon} or lack multi-illumination conditions necessary to jointly evaluate geometry and appearance under illumination shifts~\cite{UrbanScene3D,xiong2024gauu,xiong2024gauu_gauuscenev2}. As a result, few works have directly addressed this challenge due to the absence of datasets and evaluation protocols needed to jointly study illumination-robust geometry, novel-view synthesis, and inverse rendering from an aerial perspective.

To this end, we present \textbf{Skylume}, a large-scale, real-world UAV oblique dataset designed for illumination-robust 3D modeling. Our dataset covers 10 urban regions with five-direction imagery (four oblique views plus nadir) captured in the morning, at noon, and in the evening. Each region includes per-scene LiDAR scans and high-precision 3D ground-truth geometry, and all time slots are precisely aligned within a unified coordinate system. {We also provide high-quality ground-truth annotations for depth, surface normals, reconstruction quality, and novel-view synthesis evaluation. Additionally, we introduce an inverse-rendering track with the Temporal Consistency Coefficient (TCC) metric to quantify cross-time stability.} By design, Skylume transforms multi-temporal capture into a testbed: methods are challenged not only to render what was observed, but also to harmonize appearance and preserve geometry across illumination changes throughout the day.

\begin{table*}[!h]
\footnotesize
\setlength\tabcolsep{10pt}
\renewcommand\arraystretch{0.8}
\vspace{-4pt}
\caption{\textbf{Comparison of aerial datasets for
 3D reconstruction.} \textbf{SkyLume} is the only real-world dataset that supports rigorous evaluation of 3D reconstruction, novel view synthesis, and inverse rendering, while prior corpora miss one or more of these axes.}
\vspace{-8pt}
\resizebox{\linewidth}{!}{
\begin{tabular}{lllcccc}
\toprule
\textbf{Aerial Dataset} & Real-World & Lidar & Camera Type & Light & Depth/Normal & Resolution \\ \midrule
ISPRS-Bencnmark ~\cite{isprs-annals-II-3-W4-135-2015} & \textcolor{green}{\ding{51}}   & \textcolor{green}{\ding{51}} \textit{(terrestrial)} & Oblique & \textcolor{red}{\ding{55}} &\textcolor{red}{\ding{55}}   & 6000$\times$4000 \\
UrbanScene3D~\cite{lin2022us3d} & \textcolor{green}{\ding{51}}\textit{(part)} & \textcolor{green}{\ding{51}} & Oblique&\textcolor{red}{\ding{55}} & \textcolor{red}{\ding{55}}   & 5490$\times$3651 \\
Mill 19 ~\cite{meganerf}& \textcolor{green}{\ding{51}} & \textcolor{red}{\ding{55}}  & Oblique & \textcolor{red}{\ding{55}} & \textcolor{red}{\ding{55}} &  4608$\times$3456  \\

Horizon-GS \cite{jiang2025horizon}& \textcolor{green}{\ding{51}}\textit{(part)} & \textcolor{red}{\ding{55}} & \textbf{Oblique+Nadir} & \textcolor{red}{\ding{55}} & \textcolor{red}{\ding{55}}   & 1600$\times$1066 \\
GauU-Scene \cite{xiong2024gauu}& \textcolor{green}{\ding{51}} & \textcolor{green}{\ding{51}} & \textbf{Oblique+Nadir} &\textcolor{red}{\ding{55}} & \textcolor{red}{\ding{55}}  & 5468$\times$3636 \\
UAVScenes \cite{wang2025uavscenes}& \textcolor{green}{\ding{51}} & \textcolor{green}{\ding{51}}  & Nadir & \textcolor{red}{\ding{55}} &\textit{Depth} & N/A   \\

Matrix City \cite{li2023matrixcity}& \textcolor{red}{\ding{55}} & \textcolor{red}{\ding{55}} \textit{(from depth)} & \textbf{Oblique+Nadir} & \textcolor{green}{\ding{51}} & \textbf{Depth+Normal}   & 1920$\times$1080 \\

\midrule
\textbf{SkyLume (ours)} & \textcolor{green}{\ding{51}} & \textcolor{green}{\ding{51}} & \textbf{Oblique+Nadir} & \textcolor{green}{\ding{51}} & \textbf{Depth+Normal} &  \textbf{6252$\times$4168}\\
\bottomrule
\end{tabular}}

\label{tab:dataset_comp}
\end{table*}


Our benchmark study highlights the value of \textbf{Skylume} for illumination-robust urban scene modeling. Evaluating representative 3DGS-based methods under that maintaining stability under varying illumination at scale remains a significant challenge: (1) Illumination shifts within the same region induce substantial variations in rendering quality, particularly for weakly textured façades, glass boundaries, and water-adjacent surfaces, leading to divergent; (2) Geometry is often corrupted by lighting effects, with cast shadows and moving penumbras frequently overfitted as solid structures, thereby biasing depth and surface normal estimates and introducing spurious geometry; and (3) Inverse rendering methods at city-scale remain fragile, as the estimated albedo retains time-dependent shadows and exposure variations, failing to achieve cross-time consistency. Together, these findings highlight the need for methods that explicitly separate light from material properties, incorporate advanced techniques for handling dynamic illumination, and ensure robust cross-time consistency in large-scale 3D reconstruction tasks.

The contributions are summarized as follows:
\begin{itemize}
    \item We present \textbf{Skylume}, a multi-temporal UAV oblique dataset targeting illumination diversity for geometry, NVS, and inverse rendering. To our knowledge, skyLume is the first real-world aerial dataset covering 10 urban regions with more than 100K images captured under different illumination conditions, paired with high-precision 3D ground-truth geometry and 6-DoF poses.
    \item We provide standardized splits and evaluation protocols for cross-time rendering and geometry evaluation, and introduce the Temporal Consistency Coefficient (TCC) metric to quantify cross-time albedo stability.
    \item We benchmark representative 3DGS variants across varying illumination conditions and reveal three key challenges in rendering quality, geometry extraction, and albedo inconsistency under illumination changes.
\end{itemize}


\section{Related Work}\label{sec:related work}

\subsection{3D Gaussian Splatting and Beyond}\label{sec:3dgs-related-work}
3D Gaussian Splatting (3DGS) \cite{kerbl20233d} represents a scene as an explicit set of anisotropic Gaussians and enables real-time, visibility-aware rendering with competitive fidelity. Since its introduction, a rapidly growing literature has extended 3DGS in terms of its applications, core algorithms, and compatibility:
\textbf{(a) Rendering-centric 3DGS} aim for stable and real-time NVS. Key advances include anti-aliasing \cite{yu2024mipsplatting}, improving densification \cite{li2024mvgsplatting, lu2024scaffold, kheradmand202433dgsmcmc, fang2024minisplattingrepresentingscenesconstrained}, improving visibility via deferred or reflective \cite{zhang2025ref-gs, yao2024ref-gaussian,xie2025envgs}, appearance decoupling for view-dependent effects \cite{xie2025envgs, yang2024specgaussiananisotropicviewdependentappearance}, and LoD or region-aware scheduling to keep memory bounded \cite{lu2024scaffold,seo2024flod,ren2024octree}. On top of these foundations, large-scene variants mainly add partitioning and distributed optimization (with hierarchical/LoD training) \cite{liu2024citygaussian,liu2024citygaussianv2efficientgeometricallyaccurate,vastgaussian,yuchen2024dogaussian,chen2024gigags,ulsrgs}, boosting throughput while preserving fidelity.
\textbf{(b) Geometry-aware 3DGS} steers Gaussians toward surfaces to obtain editable, watertight geometry. Core ideas include surface-aligned Gaussians with fast mesh extraction and joint refinement \cite{guedon2023sugar}, and surfel-like primitives that stabilize normals and coverage \cite{huang20242dgs,wolf2024surface_3dgs,chen2024pgsr}. Hybrids with depth/normal-aware training \cite{zhang2024rade,turkulainen2024dnsplatter,Yu2024GOF} further improve mesh quality and downstream editability.
\textbf{(c) Inverse-rendering 3DGS} augments Gaussians with explicit shading, materials, and illumination to enable relighting and editing. Representative directions include (i) joint recovery of geometry, BRDF, and environment lighting \cite{liang2024gs-ir}, and (ii) deferred/reflective pipelines that better handle speculars and visibility \cite{ye20243dgsdr,zhang2025ref-gs,jiang2024gaussianshader}; some further integrate differentiable ray tracing and environment Gaussians for real-time, view-dependent effects \cite{yao2024ref-gaussian,xie2025envgs}. 

Despite their strong performance, these methods are primarily designed for single-session captures and lack the ability to ensure multi-temporal consistency. This limitation is precisely the gap our benchmark addresses.

\subsection{UAV Datasets for 3D Reconstruction}
UAV datasets for 3D modeling fall into two buckets: \textbf{(a) Synthetic city-scale} corpora offer full ground truth, accurate poses, and controllable illumination/weather, which are ideal for ablations but carry domain gaps to real flights \cite{li2023matrixcity,zhao2024aerialgo,Fonder2019MidAir,vuong2025aerialmegadepth}. These methods are built on game engines, provides aerial–street views with ground-truth cameras and flexible light/weather control for city-scale reconstruction \cite{tartanair2020iros, rizzoli2023syndrone,jiang2025horizon}. 
\textbf{(b) Real-world captured} corpora better reflect deployment but rarely include cross-time revisits under controlled capture, complicating evaluation of illumination robustness. Some mix large synthetic cities with a limited set of real scenes, where capture conditions within a scene can vary \cite{UrbanScene3D, meganerf,isprs-annals-II-3-W4-135-2015,jiang2025horizon}. Some covers large urban areas with RGB and LiDAR and emphasizes reconstruction under observed lighting \cite{xiong2024gauu}. Some contributes frame-wise semantic labels for images and LiDAR, accurate 6-DoF poses, and reconstructed maps, broadening multi-modal evaluation but focusing primarily on perception rather than systematic multi-temporal oblique photogrammetry \cite{wang2025uavscenes,li2023sgloc,yan2022crossloc,wu2024uavd4l,nguyen2022ntu_viral,dhrafani2024firestereo}. 

As shown in Tab.~\ref{tab:dataset_comp}, the aim of \textbf{SkyLume} is to address these above discussed gaps with a large-scale, real-world UAV \emph{oblique} benchmark. It supports reproducible protocols for cross-time material/illumination consistency and façade-rich reconstruction at scale.

\section{The SkyLume Dataset}\label{sec:skylume}

\begin{figure*}[t]
    \centering
    \includegraphics[width=0.9\linewidth]{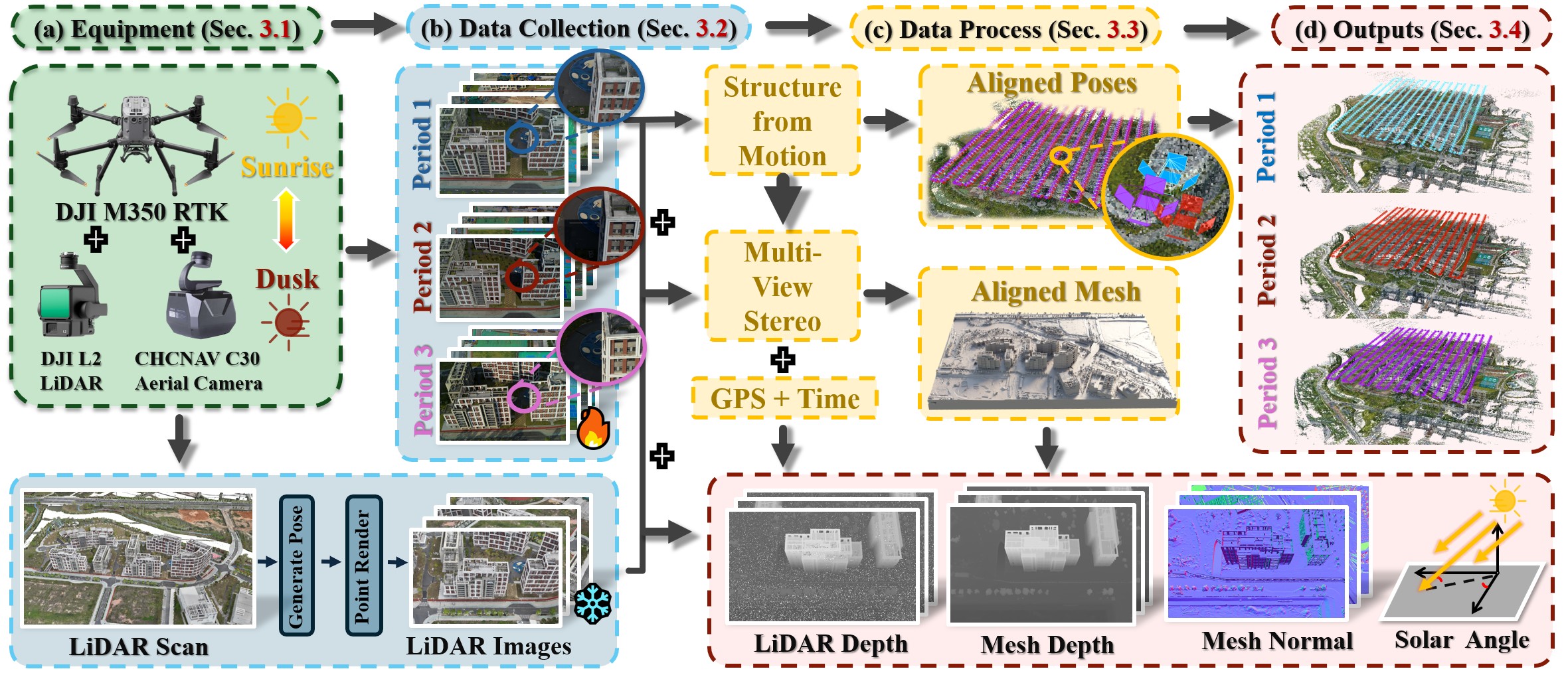}
    \vspace{-8pt}
   \caption{\textbf{Dataset collection and processing pipeline.} (a) A survey-grade UAV stack flies the same RTK-guided route at three times of day to capture five-direction 6K imagery and LiDAR. (b) A unified LiDAR-guided SfM registration includes three periods and refines poses by point-rendering LiDAR into the cameras. (c) A LiDAR-guided MVS to produce an high-quality aligned ground-truth geometry. (d) We release per-period split SfM packages and per-frame LiDAR depth, mesh depth, mesh normals, and solar geometry.}
    \vspace{-8pt}
    \label{fig:pipeline}
\end{figure*}

The \textbf{Skylume} dataset is designed to address the limitations of existing datasets by providing comprehensive and high-quality data to support robust performance under varying illumination conditions. Fig.~\ref{fig:pipeline} illustrates the overall pipeline. Section \ref{sec:equip} details the data acquisition setup, while Section \ref{sec:collection} provides an overview of the data collection process. Section \ref{sec:preprocess} outlines the data processing methods, and Section \ref{sec:output} presents the resulting high-quality data.  Detailed per scene statics can be found at supplementary material.

\subsection{Equipment}\label{sec:equip}
\begin{figure}[t]
    \centering
    \includegraphics[width=0.9\linewidth]{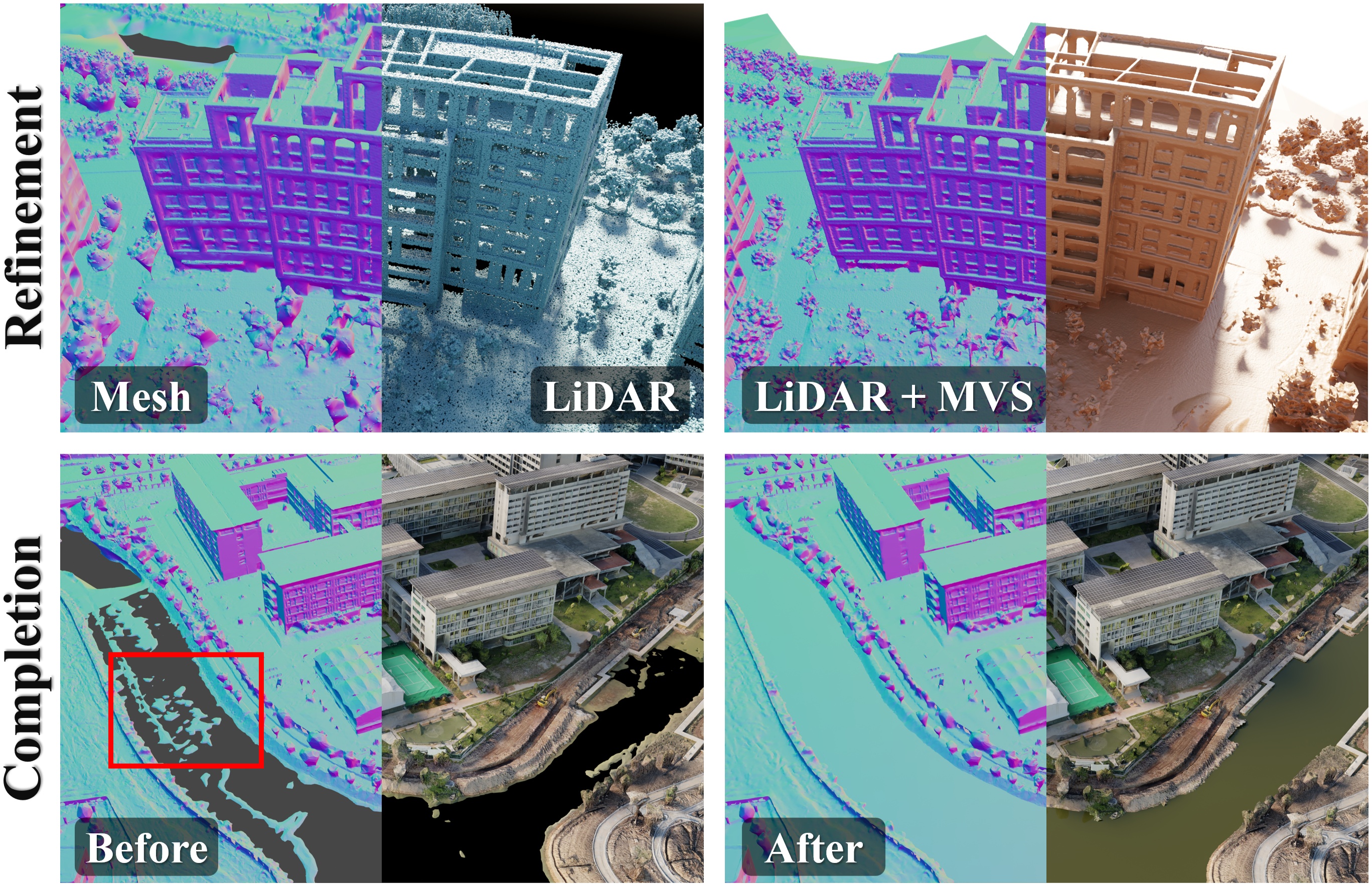}
    \vspace{-4pt}
    \caption{\textbf{Geometry post-processing.} We first build geometry ground truth via LiDAR-guided MVS. For reflective area such as river and lake, we manually repair water surfaces to correct MVS failures and ensure geometric continuity.}
    \vspace{-0pt}
    \label{fig:postprocessing}
\end{figure}
As shown in Fig.~\ref{fig:pipeline}, we employ a survey-grade setup designed for stable flights, high-resolution multi-view imagery, and accurate point clouds for reliable alignment and evaluation:
\textit{(a) DJI Matrice 350 RTK:} A UAV equipped with RTK-grade positioning and long endurance, enabling repeatable flights across different times of day. \textit{(b) CHCNAV C30 Aerial Oblique Camera:} A 130 MP camera featuring four $45^\circ$ oblique views and one $90^\circ$ nadir view, captured synchronously to enhance façade coverage and minimize drift. \textit{(c) DJI Zenmuse L2 LiDAR:} A frame-scanning LiDAR with 5 cm horizontal and 4 cm vertical accuracy at 150 m, ensuring precise alignment.
Detailed equipment specifications can be found in the supplementary material.

\begin{figure*}[!h]
    \centering
    \includegraphics[width=0.99\linewidth]{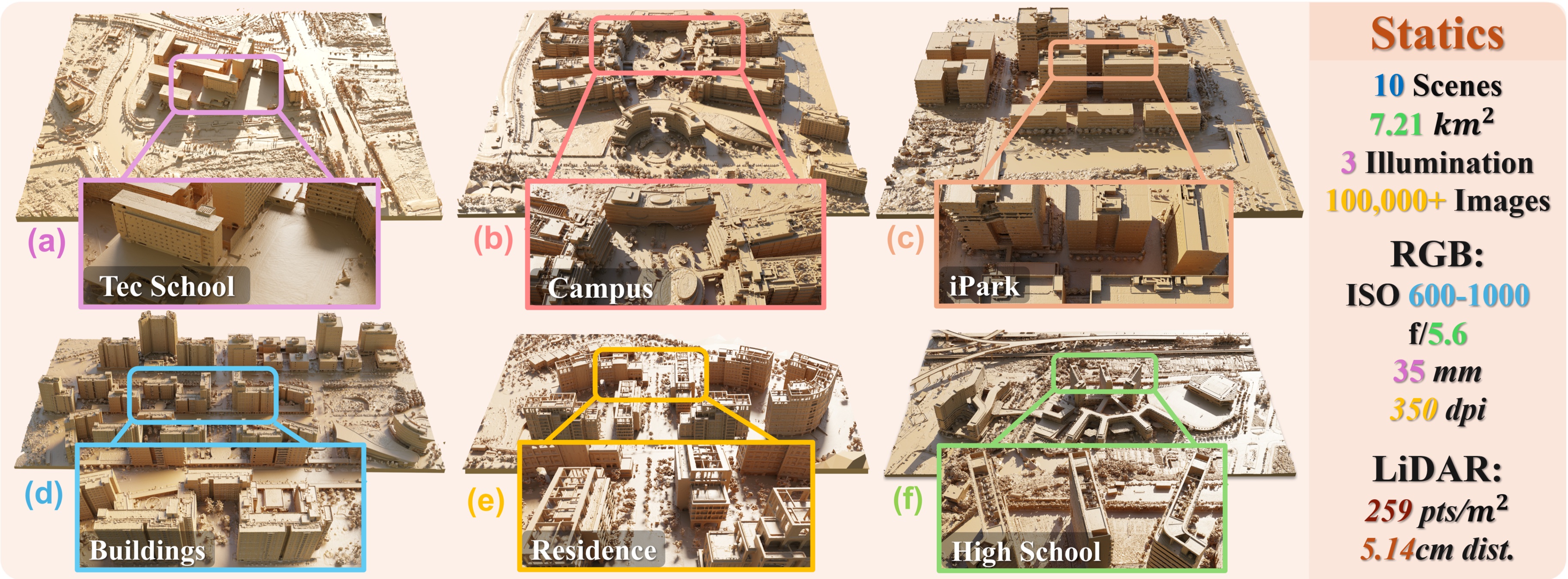}
    \vspace{-8pt}
    \caption{\textbf{Visualization of the ground-truth models.} We visualize post-processed meshes for six representative medium-scale scenes.}
    \vspace{-8pt}
    \label{fig:lidarmesh}
\end{figure*}

\subsection{Data  Collection}\label{sec:collection}
As illustrated in Fig.~\ref{fig:pipeline}, data is collected at three distinct times of day to capture varying illumination conditions: early morning (07:00–09:00), midday (11:00–13:00), and late afternoon (16:00–18:00). To ensure consistency across these lighting conditions, the same waypoint trajectory is repeated for each time slot, maintaining comparable viewpoints and coverage.

The flight plan follows standard photogrammetric practices. The forward overlap is set to 80$\%$, and the side overlap is 60$\%$. Flights are conducted at an altitude of 120 \textit{m} above ground level, with an airspeed of 8 \textit{m/s}. The CHCNAV C30 camera is triggered at 1 Hz, capturing five synchronized views to maximize façade and roof coverage while minimizing viewpoint and exposure drift. RTK corrections are applied throughout the flight, and post-differential positioning is recorded in the EXIF metadata to provide precise WGS 84 pose estimates for all images. LiDAR data is acquired separately on the following day, since it is an active measurement modality, remains unaffected by illumination variations. 

\subsection{Data Preprocessing}\label{sec:preprocess}

\textbf{Multi-temporal SfM Alignment.} To facilitate accurate multi-modal alignment, all data is projected into a common coordinate system. The LiDAR data is georeferenced in WGS 84 / UTM, while RGB images contain RTK-aided  metadata.
We use \textit{RealityScan} to register the LiDAR point cloud and generate fixed camera poses on the LiDAR point cloud. As shown in Fig.~\ref{fig:pipeline}, these fixed poses are then used as alignment anchors to render pseudo-RGB LiDAR views. Subsequently, the camera poses of all RGB images across the different time slots \textbf{are forced to align with the LiDAR poses}. A joint structure-from-motion (SfM) process is then performed across all three time slots, incorporating both the RGB images and the pseudo-RGB LiDAR renders. The RTK pose priors serve as soft constraints to initialize the camera poses, which are refined during global bundle adjustment. In areas with sparse features where residual drift may occur, we correct for any misalignment by augmenting the solution with a small set of manually selected ground control points (GCPs). A detailed report on the alignment process, including error metrics and validation, can be found in the Supplemental Materials.

\noindent\textbf{LiDAR-guided meshing.} Given the aligned multi-temporal poses, we perform surface reconstruction in \textit{RealityScan} by fusing the three-time-slot RGB imagery with the LiDAR data. The LiDAR provides a metric scaffold that regularizes depth in weak-texture and shadowed regions. As shown in Fig.~\ref{fig:postprocessing}, the final output is a high-fidelity dense reconstruction, which outperforms reconstruction using RGB images alone. 

\noindent\textbf{Geometry completion.} Rivers and ponds pose challenges for MVS due to view-dependent specularity and translucency. To address this, we estimate a physically plausible flat surface at the shoreline using LiDAR data. For each river or pond, we select four shoreline samples and compute the mean elevation $z_{mean}$ (with a sample spread of $\le 2$ cm). Surface normals are regularized to be near horizontal, and the patch is re-meshed (see Fig.~\ref{fig:postprocessing}) to ensure continuity with the surrounding geometry. 

Fig.~\ref{fig:lidarmesh} demonstrates the geometry ground-truth results after refinement and completion.

\subsection{Output Data} \label{sec:output}

To ensure seamless integration with existing pipelines and facilitate reproducible evaluation, we export each reconstruction region in the COLMAP format.

\noindent \textbf{SfM Results for Three Time Slots.} From the global camera extrinsics, we isolate the \textit{image name} corresponding to each of the three time slots, creating three distinct subsets of camera poses. Using these time-specific extrinsics, we then match and extract the corresponding camera intrinsics and sparse point cloud from the global SfM solution. 

\noindent \textbf{Per-frame Depth and Normals.} We provide two distinct modalities for depth and normal data: (a) \textit{Mesh Depth and Normals} are derived by projecting the mesh into each camera view. The resulting depth and normal maps are dense, with resolutions matching those of the original images. (b) \textit{LiDAR Depth} is obtained by re-projecting the LiDAR point cloud into the camera frames after occupancy filtering. Although this modality is sparser, it offers higher metric reliability for accurate depth estimation, as the removal of background objects prevents erroneous points from being projected.
\begin{figure}[t]
    \centering
    \vspace{-8pt}
    \includegraphics[width=1\linewidth]{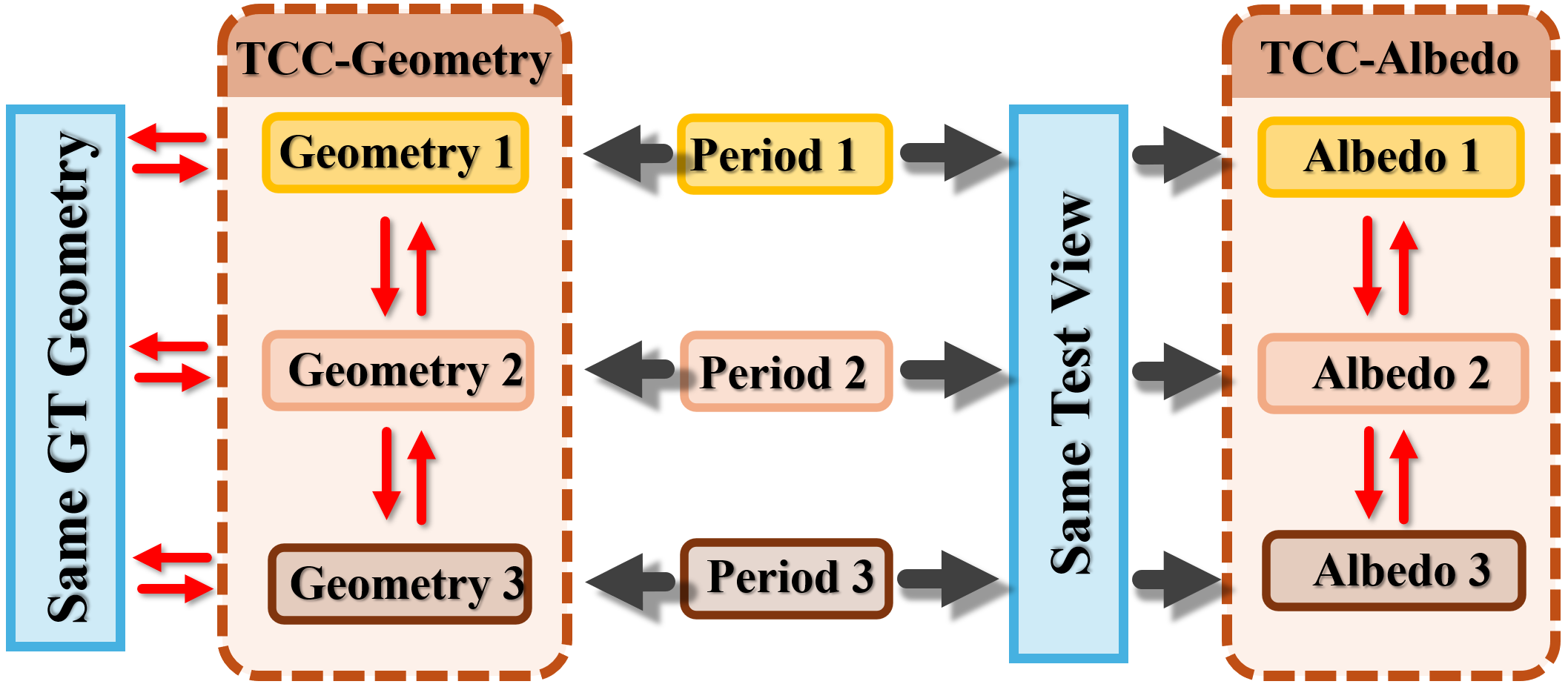}
    \vspace{-8pt}
\caption{\textbf{Temporal Consistency Coefficient (TCC) evaluation}. For inverse rendering, we render albedo from identical test viewpoints, and compute TCC-Albedo across periods.  For Geometry, We evaluate each mesh against ground-truth, and compute TCC-Geometry by averaging pairwise consistency between period meshes.}
\vspace{-8pt}
    \label{fig:eval}
\end{figure}

\begin{table*}[!htpb]
\footnotesize
    \setlength\tabcolsep{3pt}
    \centering 
    \renewcommand\arraystretch{1}
    \vspace{-8pt}
\caption{\textbf{{TCC metric} across three inverse‐rendering baselines.} We report the distribution of sub-metrics (TCC-LPIPS, TCC-SSIM, TCC-MAE) and the combined \textbf{TCC Overall} with Mean, Min, Max, and Std. All scores lie in [0,1], higher is better except Std.}
\vspace{-8pt}
     \resizebox{\textwidth}{!}{
\begin{tabular}{l|ccc|ccc|ccc|ccc}
\toprule

 & \multicolumn{3}{c|}{\textbf{Mean}$\uparrow$} & \multicolumn{3}{c|}{\textbf{Min}$\uparrow$} & \multicolumn{3}{c|}{\textbf{Max}$\uparrow$} & \multicolumn{3}{c}{\textbf{Std}$\downarrow$} \\ \midrule
\textit{methods} & \multicolumn{1}{l}{GS-IR \cite{liang2024gs-ir}} & \multicolumn{1}{l}{Ref-Gaussian \cite{yao2024ref-gaussian}} & \multicolumn{1}{l|}{Ref-GS \cite{zhang2025ref-gs}} & \multicolumn{1}{l}{GS-IR \cite{liang2024gs-ir}} & \multicolumn{1}{l}{Ref-Gaussian \cite{yao2024ref-gaussian}} & \multicolumn{1}{l|}{Ref-GS} & \multicolumn{1}{l}{GS-IR} & \multicolumn{1}{l}{Ref-Gaussian} & \multicolumn{1}{l|}{Ref-GS \cite{zhang2025ref-gs}} & \multicolumn{1}{l}{GS-IR \cite{liang2024gs-ir}} & \multicolumn{1}{l}{Ref-Gaussian \cite{yao2024ref-gaussian}} & \multicolumn{1}{l}{Ref-GS \cite{zhang2025ref-gs}} \\
\midrule
TCC-LPIPS  & 0.826 & \underline{0.866} & \textbf{0.874} & \underline{0.759} & \textbf{0.778} & 0.755 & 0.867 & \underline{0.919} & \textbf{0.979} & \textbf{0.020} & \underline{0.027} & 0.029 \\
TCC-SSIM & \underline{0.905} & 0.883 & \textbf{0.928} & \underline{0.864} & 0.832 & \textbf{0.878} & \underline{0.936} & 0.928 & \textbf{0.985} & \textbf{0.014} & 0.019 & \underline{0.017} \\
TCC-MAE & \underline{0.700} & 0.513 & \textbf{0.766} & 0.212 & \underline{0.262} & \textbf{0.587} & \underline{0.800} & 0.653 &\textbf{ 0.977} & \underline{0.074} & 0.078 & \textbf{0.063} \\
\midrule
\textbf{TCC Overall} & \underline{0.721} & 0.658 &\textbf{ 0.775} & 0.563 & \underline{0.575} & \textbf{0.622} & \underline{0.765} & 0.735 & \textbf{0.913} & \underline{0.033} & \textbf{0.026} & 0.036\\ 
\bottomrule
\end{tabular}}

\label{tab:tcc}
\end{table*}

\begin{figure*}[!h]
    \centering
    \includegraphics[width=0.9\linewidth]{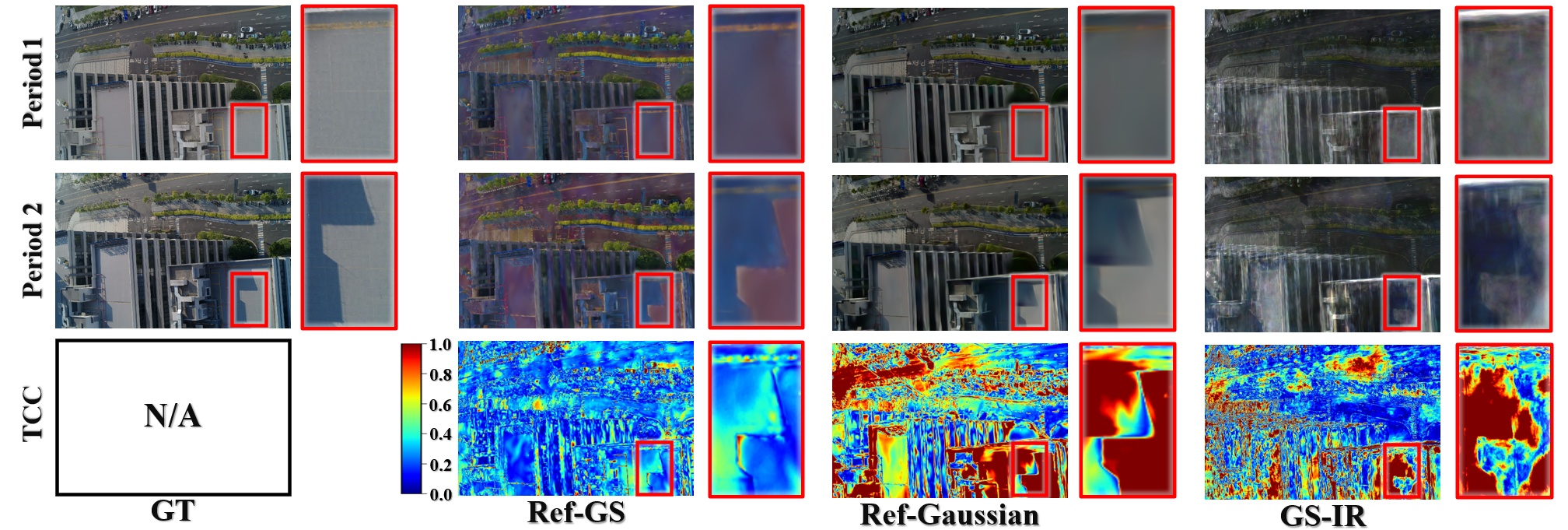}
    \vspace{-8pt}
    \caption{\textbf{Albedo and TCC visualization across two time slots.} Note that \emph{Period 2} is rendered from the viewpoints of \emph{Period 1} to fix camera pose, so differences arise solely from illumination. }
    \label{figure:alb}
    \vspace{-8pt}
\end{figure*}

\noindent\textbf{Per-frame solar geometry.} Using the capture timestamp and geolocation of each image, we provide solar elevation and solar azimuth using the NOAA Solar Geometry Calculator. We hope these annotations can support future de-shadowing, inverse rendering, and relighting studies.

\section{Benchmark Experiments}\label{sec:exp}

In this section, we evaluate three tracks on six small- and medium-scale scenes (see Fig.~\ref{fig:lidarmesh}): inverse rendering methods GS-IR \cite{liang2024gs-ir}, Ref-Gaussian \cite{yao2024ref-gaussian}, and Ref-GS \cite{zhang2025ref-gs}; geometry methods  2DGS \cite{huang20242dgs}, PGSR \cite{chen2024pgsr}, GOF \cite{Yu2024GOF}, and CitygaussianV2 \cite{liu2024citygaussianv2efficientgeometricallyaccurate}; and novel view synthesis methods 3DGS \cite{3dgs}, Abs-GS \cite{ye2024absgs}, Mip-SPlatting \cite{yu2024mipsplatting}, and Octree-GS \cite{ren2024octree}. To better evaluate the performance of each method in large-scale scenes, we conducted benchmarks on an NVIDIA RTX A800 80G GPU, applying identical adjustments to the training pipeline and parameter settings for each method. \textit{For detailed implementation information, please refer to the supplementary material.}
\subsection{Benchmark Metrics}\label{exc:metrics}
\noindent\textbf{Illumination robustness.} In real-world conditions, it is not feasible to capture accurate ground-truth albedo. Therefore, demonstrated in Fig.~\ref{fig:eval}, we propose Temporal Consistency Coefficient (\textbf{TCC}) to evaluate inverse rendering robustness by rendering albedo from the same test viewpoints across three time slots since all images are registered by a unified SfM solution. {From the global trajectory, we select} \(K\) fixed test viewpoints \(\{v_k\}_{k=1}^{K}\) using nearest pose matching and uniform spatial coverage. 
To evaluate the consistency of albedo across different illumination conditions, we use a combination of four metrics: \textit{MAE} and \textit{RMSE} assess pixel-level accuracy and color consistency, helping to ensure the albedo maintains precision across varying lighting. \textit{SSIM} measures structural similarity, evaluating how well the albedo preserves the underlying surface patterns. \textit{LPIPS}, which compares deep features of the images, captures perceptual differences and ensures that the albedo maintains visual plausibility across time slots, especially under complex lighting conditions.

{At each test viewpoint, we render the three albedo maps} \(\{A_t^{(k)}\}_{t=1}^{3}\). For each viewpoint \(v_k\), we compute the combined score \(\mathrm{TCC}^{(k)}_{\mathrm{comb}}\in[0,1]\) as: 
{\small \(\mathrm{TCC}^{(k)}_{\mathrm{comb}} =
\alpha\,\mathrm{TCC}^{(k)}_{\mathrm{MAE}}
+ \beta\,\mathrm{TCC}^{(k)}_{\mathrm{RMSE}}
+ \theta\,\mathrm{TCC}^{(k)}_{\mathrm{SSIM}}
+ \gamma\,\mathrm{TCC}^{(k)}_{\mathrm{LPIPS}}\)},
\noindent where $\alpha=0.2$, $\beta=0.2$, $\theta=0.2$, and $\gamma=0.4$. The final TCC score is the arithmetic mean over the \(K\) viewpoints:
$\mathrm{TCC}_{\mathrm{overall}}
= \frac{1}{K}\sum_{k=1}^{K}\mathrm{TCC}^{(k)}_{\mathrm{comb}}.$


\noindent\textbf{Geometry robustness.}
Demonstrated in Fig.~\ref{fig:eval}, we evaluate robustness of reconstructed geometry under changing illumination in two ways. First, we measure absolute accuracy against per scene GT geometry in Fig.~\ref{fig:lidarmesh}. Second, we measure cross-time consistency between meshes reconstructed from the three time slots. 

\noindent\textbf{Novel view synthesis.}
We additionally evaluate novel view synthesis using standard image quality metrics, reporting PSNR, SSIM~\cite{wang2004ssim}, and LPIPS~\cite{zhang2018Lpips}.
\begin{table*}[!h]
\footnotesize
    \setlength\tabcolsep{3pt}
    \centering 
    \renewcommand\arraystretch{1}
    \vspace{-4pt}
\caption{\textbf{TCC metric across four geometry baselines.} For each method we report precision (Pre), recall (Rec), and F-1 at three distance tolerances $\tau\in\{0.25,0.5,0.75\}$ m for Period 0 and Period 1. The rightmost block gives temporal consistency as the average pairwise F-1 between the period meshes at the same distance tolerances.}
 \vspace{-8pt}
     \resizebox{\textwidth}{!}{
\begin{tabular}{l|ccc|ccc|ccc|ccc|ccc|ccc|ccc}
\toprule
 & \multicolumn{9}{c|}{\textbf{Period 1}} & \multicolumn{9}{c|}{\textbf{Period 2}} & \multicolumn{3}{c}{\textbf{TCC-Geometry}} \\ \midrule
 $\tau$ (meters) & \multicolumn{3}{c|}{0.25} & \multicolumn{3}{c|}{0.5} & \multicolumn{3}{c|}{0.75} & \multicolumn{3}{c|}{0.25} & \multicolumn{3}{c|}{0.5} & \multicolumn{3}{c|}{0.75} & 0.25  & 0.5 & 0.75 \\ \midrule
 \textit{metrics} & Pre$\uparrow$ & Rec$\uparrow$ & \multicolumn{1}{c|}{F-1}$\uparrow$ & Pre$\uparrow$ & Rec$\uparrow$ & \multicolumn{1}{c|}{F-1} & Pre$\uparrow$ & Rec$\uparrow$ & F-1$\uparrow$ & Pre$\uparrow$ & Rec$\uparrow$ & \multicolumn{1}{c|}{F-1} & Pre$\uparrow$ & Rec$\uparrow$ & \multicolumn{1}{c|}{F-1}$\uparrow$ & Pre$\uparrow$ & Rec$\uparrow$ & F-1$\uparrow$ & F-1$\uparrow$ & F-1$\uparrow$ & F-1$\uparrow$ \\ \midrule
 
\textbf{2DGS \cite{huang20242dgs}} & \underline{0.203} & \underline{0.259} & \underline{0.227} & \underline{0.462} & \underline{0.562} & \underline{0.508} & 0.663 & 0.750 & 0.704 & \underline{0.199} & \textbf{0.278} & \underline{0.232} & \underline{0.486} & \textbf{0.639} & \underline{0.552} & 0.672 & \textbf{0.811} & \underline{0.735} & \textbf{0.570} & \underline{0.675} & \underline{0.750} \\

\textbf{PGSR \cite{chen2024pgsr}} & 0.136 & 0.171 & 0.152 & 0.382 & 0.485 & 0.428 & 0.680 & \underline{0.764} & \underline{0.719} & 0.150 & 0.198 & 0.171 & 0.368 & 0.489 & 0.420 & 0.671 & \underline{0.778} & 0.720 & \underline{0.554} & 0.662 & 0.749 \\ 

\textbf{GOF \cite{Yu2024GOF}} & 0.106 & 0.101 & 0.103 & 0.417 & 0.336 & 0.372 & \underline{0.775} & 0.594 & 0.672 & 0.120 & 0.113 & 0.116 & 0.455 & 0.357 & 0.400 & \textbf{0.810} & 0.599 & 0.689 & 0.431 & 0.610 & 0.712 \\
\textbf{CityGaussianV2 \cite{liu2024citygaussianv2efficientgeometricallyaccurate}} &  \textbf{0.315} & \textbf{0.270} & \textbf{0.291} & \textbf{0.618} & \textbf{0.603} & \textbf{0.610} & \textbf{0.782} & \textbf{0.835} & \textbf{0.808} & \textbf{0.256} & \underline{0.245} & \textbf{0.250} & \textbf{0.563} & \underline{0.569} & \textbf{0.566} & \underline{0.797} & 0.746 & \textbf{0.770} & 0.547 & \textbf{0.719} & \textbf{0.790} \\ \bottomrule
\end{tabular}}
 
\label{tab:geo}
\end{table*}
\begin{figure*}[!ht]
    \centering
    \includegraphics[width=1\linewidth]{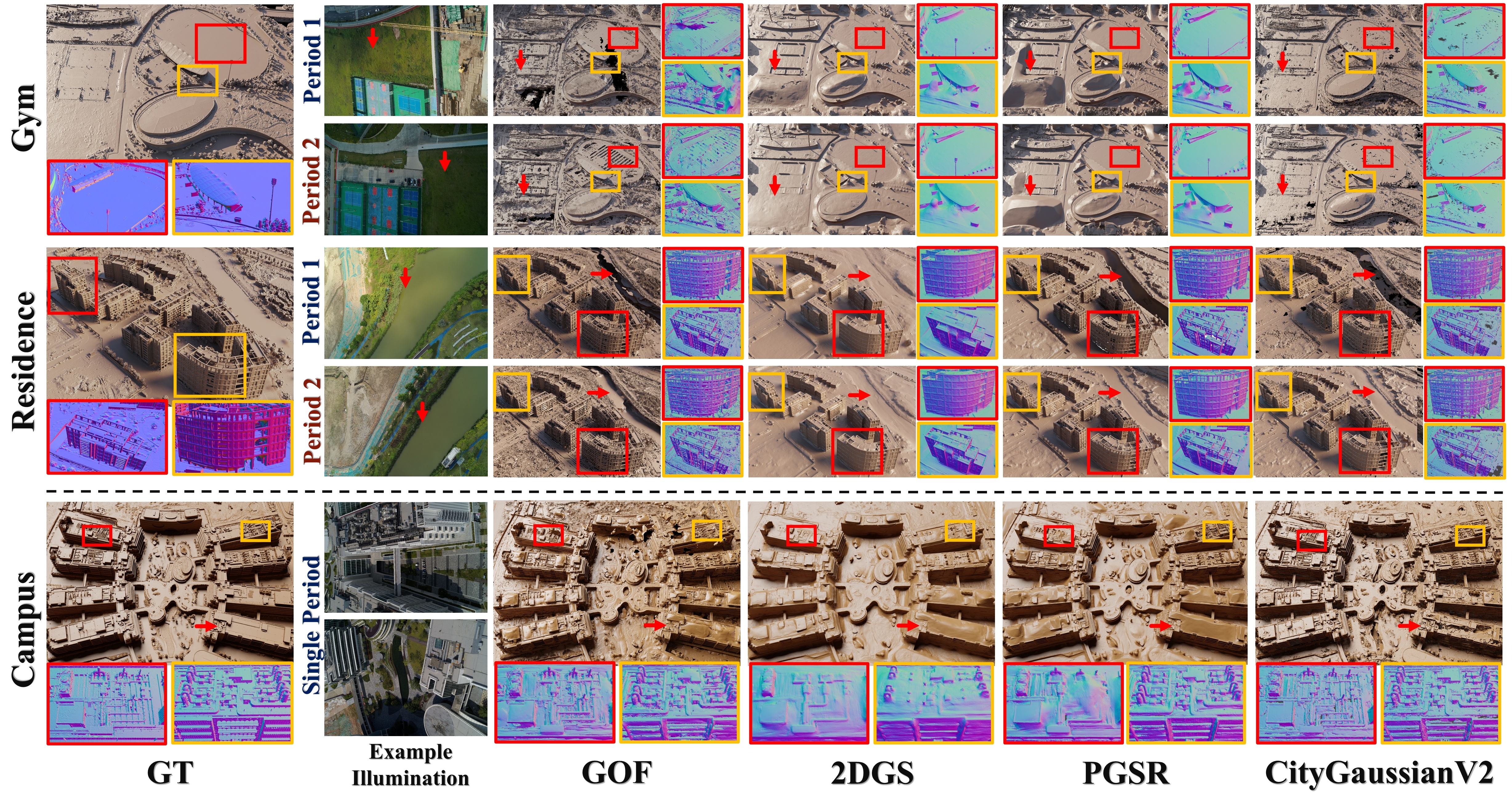}
    
    \vspace{-8pt}
    \caption{\textbf{Geometry visualization.} \textbf{Top}: Geometry for \emph{Gym} and \emph{Residence}, \textcolor{red}{red} and \textcolor{orange}{yellow} frames are the zoomed-in surface normals. \textbf{Bottom}: Single Period 1 mesh comparison. Under the sunlit \emph{Period 1}, all methods exhibit holes and breakups.}
    \vspace{-8pt}
    \label{fig:geo}
\end{figure*}

\subsection{Cross-Time Albedo Evaluation}

We assess temporal albedo consistency across three inverse-rendering baselines using our TCC metric, which rewards cross-time stability rather than single-slot fidelity. Tab.~\ref{tab:tcc} shows that albedo estimates drift under illumination changes even at matched viewpoints, with large dispersion across scenes in the Min, Max, and Std columns. The perceptual branch TCC-MAE is frequently the limiting factor while TCC-SSIM remains comparatively stable, indicating that \textbf{structure is preserved more often than appearance}. Fig.~\ref{figure:alb} corroborates these trends on two regions with strong lighting shifts where all methods retain shadow imprinting.  As a result, the proposed TCC turns multi-temporal robustness into a measurable target, enabling fair and reproducible comparison of illumination-robust inverse rendering.

\subsection{Cross-Time Geometry Evaluation}
We evaluate geometry under illumination change by reporting precision, recall, and F-1 against GT for each time slot, and by measuring cross-time consistency via the average pairwise consistency F-1 score between the the meshes in Tab.~\ref{tab:geo}. Two lighting regimes are highlighted in Fig.~\ref{fig:geo}: a sunlit case with strong, moving shadows and color shifts, and an overcast case dominated by diffuse irradiance. Geometry extracted in diffuse conditions is the most stable, while direct sunlight degrades both absolute accuracy and cross-time agreement. {Methods that using alpha-blending depth for TSDF-fusion \cite{werner2014tsdf} (2DGS \cite{huang20242dgs} and PGSR \cite{chen2024pgsr}) preserve mesh topology more consistently across slots but lack geometry details.} Approaches that rely on opacity (GOF \cite{Yu2024GOF} and CityGaussianV2 \cite{liu2024citygaussianv2efficientgeometricallyaccurate}) can reach high per-slot F-1 score under favorable lighting but are brittle under hard shadows. Fig.~\ref{fig:geo} reveals holes and breakups in sunlight-swept regions such as lawn and water-body. 
\begin{figure}[!h]
    \centering
    \includegraphics[width=1\linewidth]{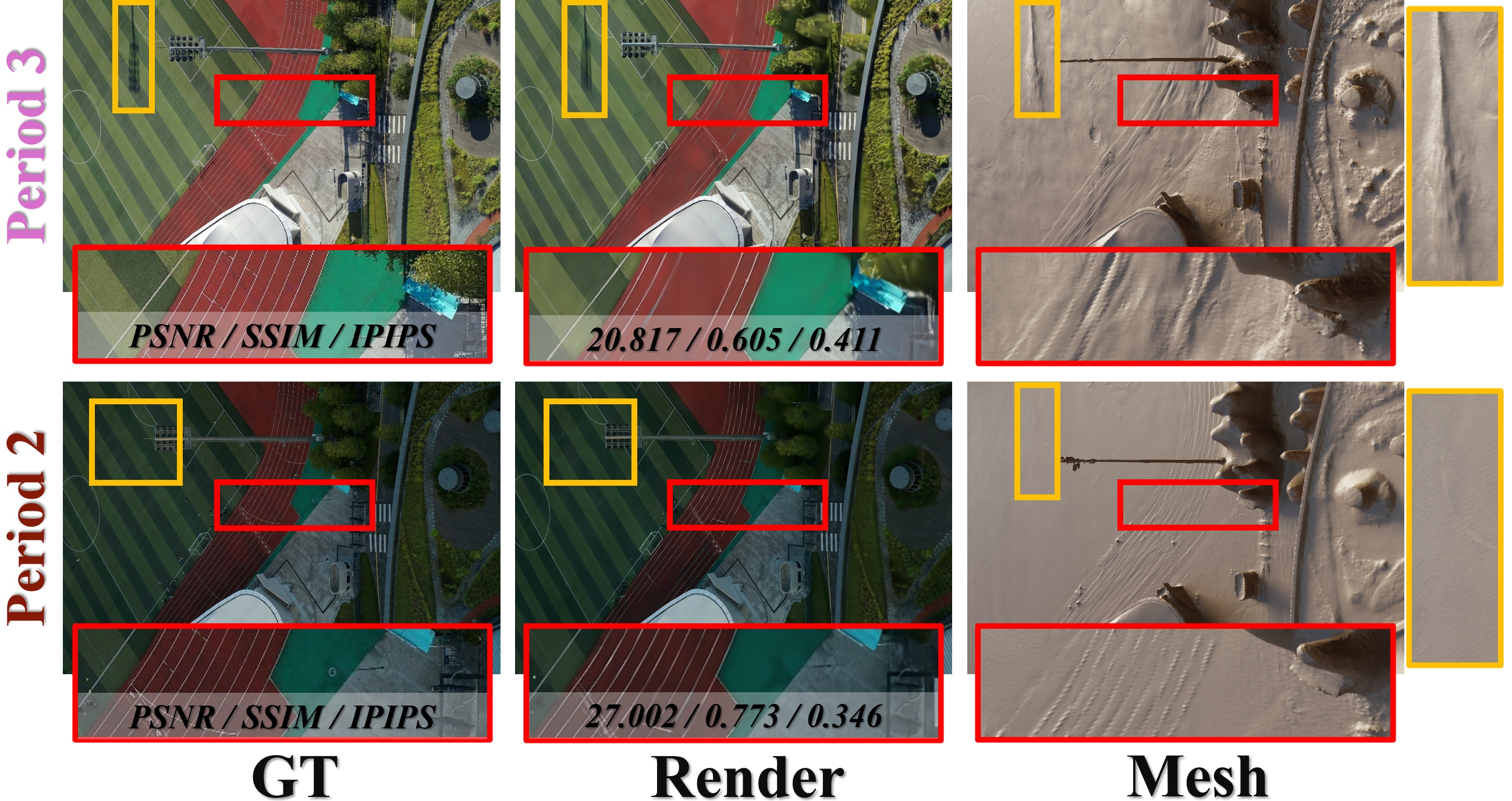}
    
    \vspace{-8pt}
    \caption{\textbf{Additional Visualization of the Impact of Strong Illumination.} Cast shadows break multi-view photometric consistency, yielding shadow-as-geometry artifacts. In reflective planes, view-dependent appearance leads to blurred textures and detail loss in NVS.}
    \label{fig:geo_nvs}
    \vspace{-8pt}
\end{figure}

\subsection{Novel View Synthesis Evaluation}

\begin{table*}[!ht]
\footnotesize
    \setlength\tabcolsep{3pt}
    \centering 
    \renewcommand\arraystretch{1}
    \vspace{-8pt}
\caption{\textbf{Novel view synthesis under \textit{Period 1} (sunlit) reconstruction.} We report SSIM, PSNR, and LPIPS across six UAV scenes for four NVS-oriented methods and three geometry-capable pipelines. "-" indicates unavailable results due to GPU out-of-memory.}
\vspace{-8pt}
     \resizebox{\textwidth}{!}{
\begin{tabular}{l|ccc|ccc|ccc|ccc|ccc|ccc}
\toprule
 & \multicolumn{3}{c|}{\textbf{GYM}} & \multicolumn{3}{c|}{\textbf{iPark}} & \multicolumn{3}{c|}{\textbf{Tec School}} & \multicolumn{3}{c|}{\textbf{Staff residence}} & \multicolumn{3}{c|}{\textbf{Campus}} & \multicolumn{3}{c}{\textbf{High School}} \\ \midrule
\textit{metrics} & SSIM$\uparrow$ & PSNR$\uparrow$ & LPIPS$\downarrow$ & SSIM$\uparrow$ & PSNR$\uparrow$ & LPIPS$\downarrow$ & SSIM$\uparrow$ & PSNR$\uparrow$ & LPIPS$\downarrow$ & SSIM$\uparrow$ & PSNR$\uparrow$ & LPIPS$\downarrow$ & SSIM$\uparrow$ & PSNR$\uparrow$ & LPIPS$\downarrow$ & SSIM$\uparrow$ & PSNR$\uparrow$ & LPIPS$\downarrow$ \\ \midrule
\textbf{3DGS} & \underline{0.672} & \underline{21.144} & \underline{0.318} & \underline{0.707} & \textbf{23.250} & \underline{0.355} & \underline{0.631} & \underline{21.652} & 0.384 & \underline{0.670} & \underline{23.099} & \underline{0.353} & 0.617 & \textbf{21.507} & \underline{0.409} & \underline{0.626} & \underline{20.570} & \underline{0.389} \\
\textbf{Abs-GS} & \textbf{0.681} & \textbf{21.183} & \textbf{0.305} & \textbf{0.711} & 22.958 & \textbf{0.335} & \textbf{0.639} & 21.523 & \underline{0.375} & \textbf{0.676} & \textbf{23.125} & \textbf{0.352} & \textbf{0.633} & \underline{21.477} & \textbf{0.390} & \textbf{0.634} & {20.413} & \textbf{0.382} \\
\textbf{Mip-Splatting} & 0.669 & 21.131 & 0.325 & 0.701 & 22.708 & 0.362 & 0.621 & 21.364 & 0.398 & 0.667 & 23.006 & 0.360 & 0.619 & 20.015 & 0.427 & - & - & - \\
\textbf{Octree-GS} & 0.648 & 20.931 & 0.336 & 0.697 & 22.681 & 0.358 & 0.607 & \textbf{21.765} & \textbf{0.360} & 0.659 & 23.094 & 0.366 & \underline{0.624} & 21.432 & 0.425 & 0.624 & \textbf{20.618} & \textbf{0.380} \\ \midrule
\textbf{GOF} & 0.626 & 20.830 & 0.360 & 0.681 & \underline{23.172} & 0.382 & 0.619 & 21.224 & 0.401 & 0.619 & 22.513 & 0.401 & 0.505 & 20.047 & 0.524 & - & - & - \\
\textbf{2DGS} & 0.583 & 20.350 & 0.425 & 0.648 & 22.722 & 0.440 & 0.549 & 21.103 & 0.489 & 0.585 & 21.891 & 0.458 & 0.507 & 20.518 & 0.539 & 0.497 & 19.531 & 0.539 \\
\textbf{PGSR} & 0.580 & 20.110 & 0.424 & 0.622 & 21.953 & 0.467 & 0.537 & 20.926 & 0.500 & 0.591 & 22.094 & 0.426 & 0.514 & 20.572 & 0.530 & 0.490 & 19.398 & 0.547 \\
\bottomrule
\end{tabular}}

\label{tab:nvs}
\end{table*}

\begin{figure*}[!ht]
    \centering
    \includegraphics[width=1\linewidth]{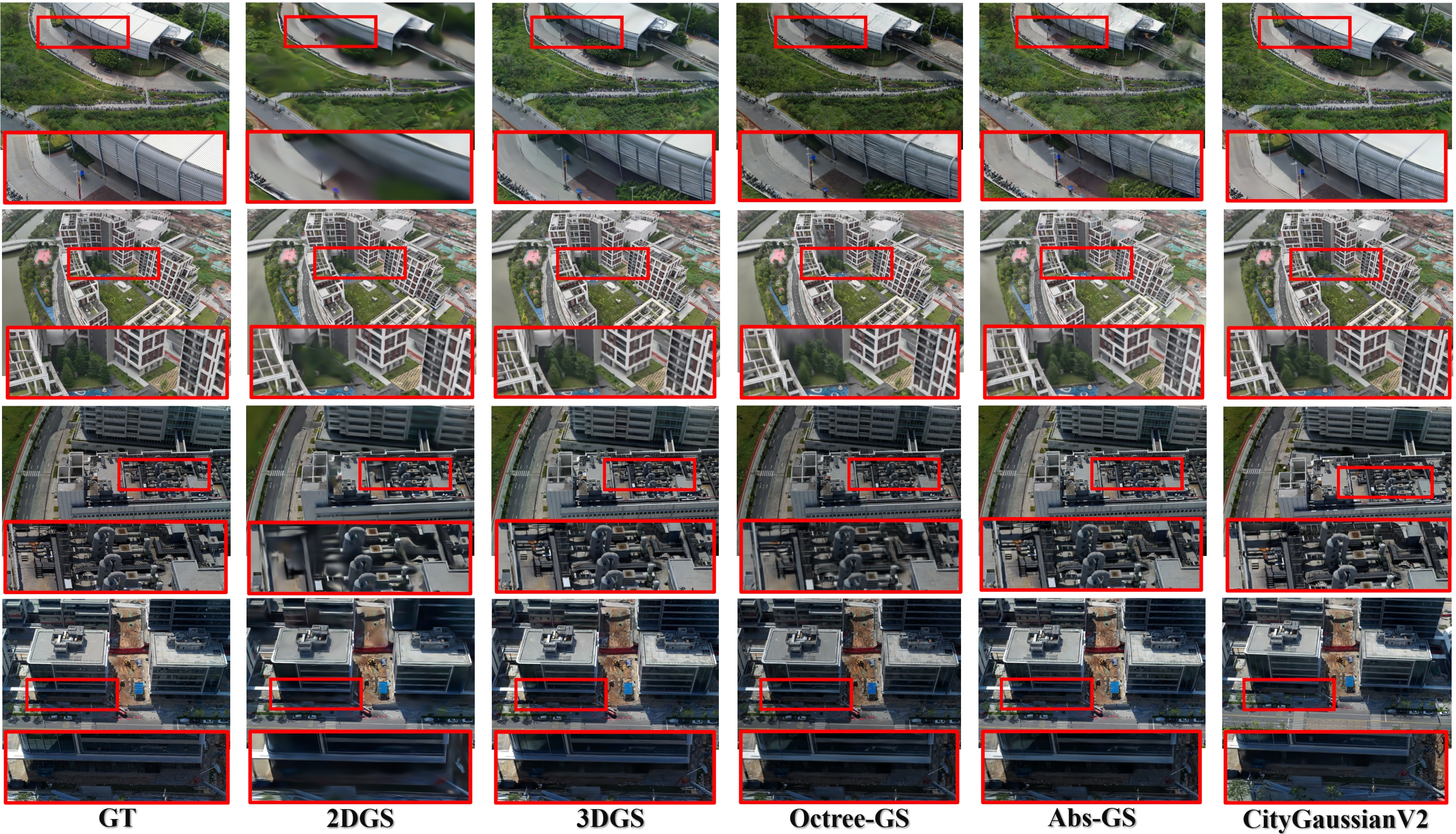}
    \vspace{-10pt}
   \caption{\textbf{Visualization of novel view synthesis under \textit{Period 1} (sunlit).} We visualize renderings from shadow-dominated regions to highlight some baselines lost details.}

    \vspace{-10pt}
    \label{fig:nvs}
\end{figure*}

Under strong lighting conditions, we evaluate four NVS-oriented methods and three geometry-oriented methods. { As summarized in Tab.~\ref{tab:nvs}, the trend is consistent across scenes: geometry-oriented pipelines incur a noticeable loss in photometric fidelity and therefore lag in rendering quality.} Within the NVS group, Abs-GS \cite{ye2024absgs} is the most reliable under UAV oblique viewpoints, and Octree-GS \cite{ren2024octree} recovers fine structures. By contrast, 2DGS \cite{huang20242dgs} produces blurred renderings in all four representative scenes, as illustrated in Fig.~\ref{fig:nvs}. In Fig.~\ref{fig:geo_nvs}, we can observe that strong sunlight introduces shadow boundaries and exposure drift, which amplify cross-view color inconsistencies, degrade local detail, and widen metric differences. 

\section{Conclusion and Future Work}\label{conclusion}
In this paper, we introduce \textbf{SkyLume}, a large-scale UAV dataset that couples high-resolution imagery captured under three different illumination conditions. The release includes unified SfM resolution, per-frame depth and normals, and solar geometry.  {Central to the dataset is the Temporal Consistency Coefficient metric, which evaluates cross-time stability rather than single-slot fidelity.}
Experiments findings point to concrete directions: future research should integrate cross-time, shadow-aware priors, and jointly reason about illumination, material, and geometry at city scale. 

\noindent\textbf{Future Work.} {We plan to further expand SkyLume with additional scenes, weather conditions, and sensing modalities to enhance robustness and support tasks such as relighting and editing.} In addition, we plan to add multimodal annotations, such as semantic segmentation and infrared imagery, to enable cross-modal fusion for illumination-robust reconstruction. Our goal is to transform multi-temporal UAV capture from a confounding factor into a powerful tool for stable inverse rendering, reliable geometry reconstruction, and high-fidelity novel view synthesis in real urban environments.


{
    \small
    \bibliographystyle{ieeenat_fullname}
    \bibliography{main}
}

\clearpage
\setcounter{page}{1}
\maketitlesupplementary

\section{Dataset Details}
\label{sec:dataset_details}
\subsection{Equipment Details}
\begin{table}[!h]
\caption{Specifications of the UAV platform, oblique camera, and LiDAR payload used in our data acquisition.}
\label{tab:equipments1}
\resizebox{\columnwidth}{!}{%
\begin{tabular}{ccl}
\toprule
\textbf{Device} &
  \textbf{Type} &
  \multicolumn{1}{c}{\textbf{Key parameters}} \\
\midrule
\textbf{DJI M350 RTK} &
  UAV platform &
  \begin{tabular}[c]{@{}l@{}}\textbf{1. Max flight time:} 55 min (no payload); \\ \textbf{2. Max payload:} 2.7 kg;   \\ \textbf{3. GNSS: GPS/BeiDou;}  \\ \textbf{4. Sensing: 6-direction.}\end{tabular} \\ 
  \midrule
\textbf{CHCNAV C30} &
  Oblique camera &
  \begin{tabular}[c]{@{}l@{}}\textbf{1. Total resolution:} 130 MP (26 MP $\times$ 5); \\\textbf{2. Image size:} $6252\times4168$ (3:2); \\ \textbf{3. Lenses:} 1$\times$$\sim$nadir $90^\circ$ + 4$\times$$\sim$oblique $45^\circ$; \\ \textbf{4. Focal lengths:} 25 mm / 35 mm; \\ \textbf{5. Minimum capture interval:} 0.8 s;  \\ \textbf{6. Weight:} 605 g; \\ \textbf{7. Dimensions:} $110\times108\times85$ mm.\end{tabular} \\  \midrule
\textbf{DJI L2} &
  LiDAR &
  \begin{tabular}[c]{@{}l@{}}\textbf{1. LiDAR range:} 450 m (50\% reflectivity); \\ \textbf{2. Point rate:} 1.2M pts/s (multi-return); \\ \textbf{3. Accuracy:} 5 cm horizontal / 4 cm vertical; \\ \textbf{4. Ranging accuracy:} 2 cm @ 150 m; \\ \textbf{5. FOV} (non-repetitive): $70^\circ\times75^\circ$; \\ \textbf{6. Max returns:} 5; \\ \textbf{7. RGB sensor:} 4/3'' CMOS, 20 MP; \\ \textbf{8. RGB FOV:} $84^\circ$; \\ \textbf{9. Weight:} 905 g;\end{tabular} \\ \bottomrule
\end{tabular}%
}
\end{table}

\paragraph{UAV Platform (DJI M350 RTK).}
We use a DJI Matrice 350 RTK as the carrier platform. It is an industrial-rank UAV with a maximum flight time of about 55 minutes without payload. The built-in RTK module provides centimeter-level positioning when used with network RTK or a base station. These properties enable us to replicate nearly identical flight trajectories and camera poses at three times of day, effectively controlling viewpoint and geometric variation so that differences across slots predominantly reflect illumination rather than changes in scene structure.

\paragraph{Oblique Imaging Sensor (CHCNAV C30).}
High-resolution multi-view imagery is provided by the CHCNAV C30 oblique camera system. The C30 integrates five synchronized lenses sharing a common APS-C sensor footprint of $23.5\times15.6$~mm. Each view has a resolution of $6252\times4168$ pixels (about 26~MP), yielding a total effective resolution of 130~MP per exposure. One lens is nadir ($90^\circ$) with a 25~mm focal length, and four lenses are oblique ($45^\circ$) with 35~mm focal length, forming a cross-shaped layout that simultaneously captures façades and roofs.
In our flights, the minimum capture interval of 1.0~s allows us to maintain high forward and side overlap at typical survey speeds, while the fixed aperture around $f/5.6$ and ISO in the range 800-1600 provide robust exposure under varying illumination. These key parameters ensure fine ground sampling, rich parallax on vertical structures, and reduced drift in following SfM/MVS and reconstruction.
\begin{figure}
    \centering
    \includegraphics[width=1\linewidth]{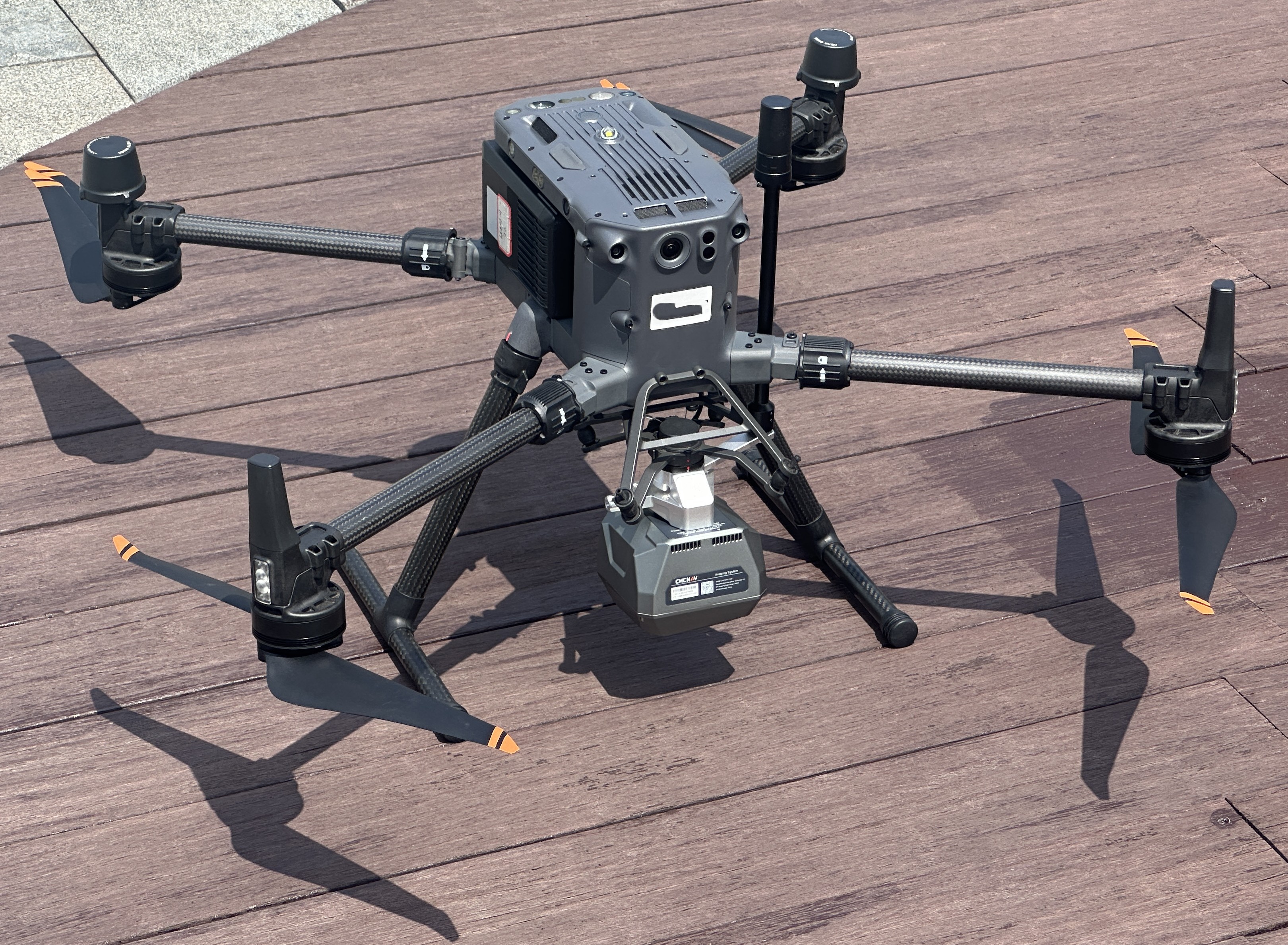}
    \vspace{-8pt}
    \caption{The DJI M350 RTK UAV platform equipped with a CHCNAV C30 oblique camera before take-off.}
    \vspace{-8pt}
    \label{fig:takeoff}
\end{figure}
\paragraph{LiDAR Ground-truth Sensor (DJI Zenmuse L2).}
To obtain accurate 3D geometry for evaluation, we mount a DJI Zenmuse L2 payload. The L2 integrates a frame LiDAR, a high-accuracy IMU, and a 4/3'' 20~MP RGB mapping camera on a 3-axis stabilized gimbal. The ranging accuracy is about 2~cm at 150~m, and under our survey configuration (relative altitude $\approx$120~m, RTK FIX, and point-cloud accuracy optimization enabled in DJI Terra), the resulting point clouds achieve approximately 5~cm horizontal and 4~cm vertical accuracy with respect to check points.
These key LiDAR characteristics (centimeter-level ranging accuracy and high-density multi-return point clouds) enable us to construct a high-fidelity reference surface via LiDAR-guided MVS reconstruction, as illustrated in Fig.~\ref{fig:postprocessing}. For highly reflective or specular regions such as rivers and lakes, we further apply manual water-surface editing on top of the L2-based model to correct MVS failures and enforce geometric continuity.

\subsection{Data Collection Details}

\begin{figure*}
    \centering
    \includegraphics[width=1\linewidth]{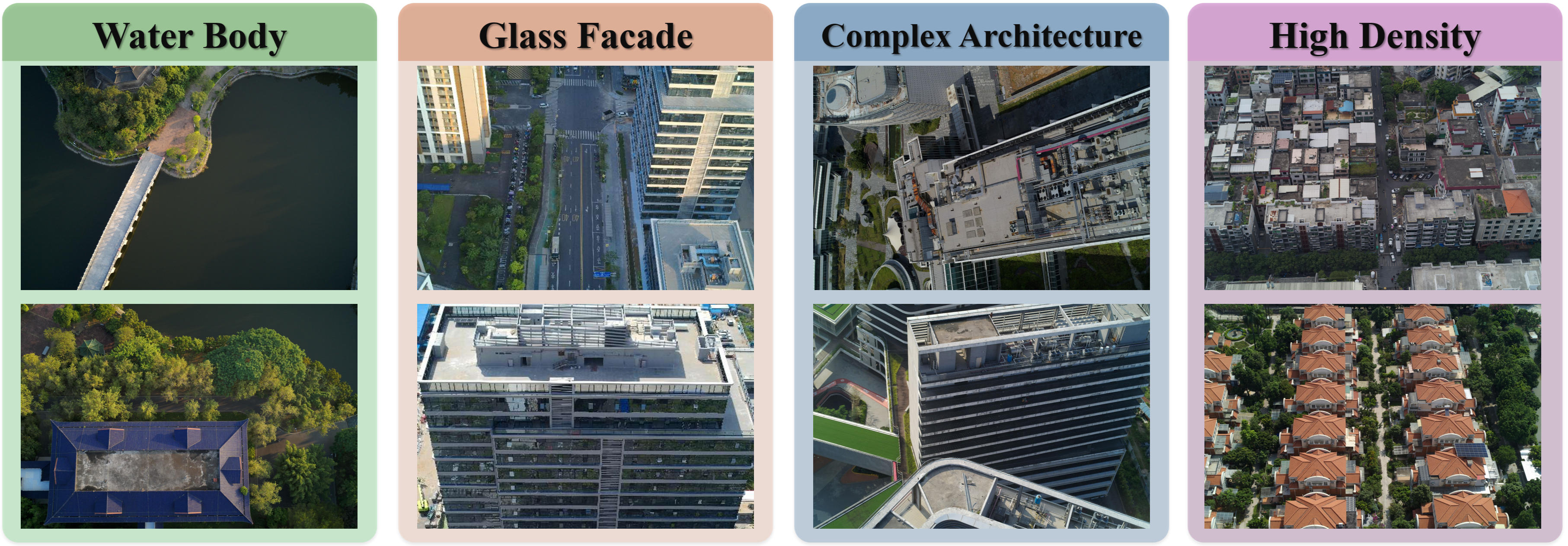}
    \vspace{-12pt}
    \caption{\textbf{Challenging cases across our dataset.}
We show representative crops for four typical difficulty patterns: (a) water bodies with strong specular reflections and weak texture, (b) glass façades with transparency and mirror-like reflections, (c) complex architectures with intricate geometry and self-occlusions, and (d) high-density urban blocks with tightly packed buildings.}
    
    \label{fig:scenestat}
\end{figure*}

\begin{table*}[!h]
\footnotesize
    \setlength\tabcolsep{8pt}
    \centering 
    \renewcommand\arraystretch{1}
\caption{\textbf{Per-scene dataset statistics.}
For each scene we report the total number of multi-view UAV images over three time slots, nominal flight height, building density, illumination configuration, and the presence of large water areas or glass façades.
Scenes are grouped into three scales: small, medium, and large, covering campus blocks to city-block-level neighborhoods.}
\vspace{-8pt}
\label{tab:data_statsic}
\resizebox{\textwidth}{!}{%
\begin{tabular}{ccccclcc}
\toprule
\textbf{Scene} &
  \textbf{Name} &
  \textbf{Total Images} &
  \textbf{Flight Height} &
  \textbf{Building Density} &
  \textbf{Illumination Type} &
  \textbf{Water Area} &
  \textbf{Glass Facade} \\ \midrule
\multirow{7}{*}{\rotatebox[origin=c]{90}{Small Scale}} &
  \textbf{Gym} &
  6185 &
  120.366 &
  Low &
  \begin{tabular}[c]{@{}l@{}}Direct Sunlight (Morning)\\ Direct Sunlight (Noon)\\ Overcast (Afternoon)\end{tabular} &
  No &
  No \\ \cline{2-8} 
\multicolumn{1}{c}{} &
  \textbf{Staff Residence} &
  7920 &
  130.018 &
  Medium &
  \begin{tabular}[c]{@{}l@{}}Direct Sunlight (Morning)\\ Partly Cloudy (Noon)\\ Overcast (Afternoon)\end{tabular} &
  Yes &
  No \\ \cline{2-8} 
\multicolumn{1}{c}{} &
  \textbf{iPark} &
  5355 &
  115.241 &
  Medium &
  \begin{tabular}[c]{@{}l@{}}Direct Sunlight (Morning)\\ Partly Cloudy (Noon)\\ Overcast (Afternoon)\end{tabular} &
  Yes &
  Yes \\ 

  \midrule
\multirow{7}{*}{\rotatebox[origin=c]{90}{Medium Scale}} &
  \textbf{Tec School} &
  7185 &
  108.481 &
  Low &
  \begin{tabular}[c]{@{}l@{}}Direct Sunlight (Morning)\\ Direct Sunlight (Noon)\\ Direct Sunlight(Afternoon)\end{tabular} &
  Yes &
  No \\ \cline{2-8} 
  
 &
  \textbf{Buildings} &
  10455 &
  129.891 &
  High &
  \begin{tabular}[c]{@{}l@{}}Direct Sunlight (Morning)\\ Direct Sunlight (Noon)\\ Overcast (Afternoon)\end{tabular} &
  No & 
  Yes \\ \cline{2-8} 
 &
  \textbf{High School} &
  10065 &
  108.015 &
  Medium &
  \begin{tabular}[c]{@{}l@{}}Direct Sunlight (Morning)\\ Partly Cloudy (Noon)\\ Direct Sunlight(Afternoon)\end{tabular} &
  Yes &
  No \\ \cline{2-8}
   
 &
  \textbf{Main Campus} &
  10410 &
  130.28 &
  Medium &
  \begin{tabular}[c]{@{}l@{}}Direct Sunlight (Morning)\\ Partly Cloudy (Noon)\\ Direct Sunlight(Afternoon)\end{tabular} &
  No &
  No \\  \midrule
\multirow{4}{*}{\rotatebox[origin=c]{90}{Large Scale}} &
  \textbf{Estate} &
  18630 &
  109.024 &
  High &
  \begin{tabular}[c]{@{}l@{}}Direct Sunlight (Morning)\\ Partly Cloudy (Noon)\\ Direct Sunlight(Afternoon)\end{tabular} &
  No &
  No \\ \cline{2-8} 
 &
  \textbf{Town} &
  12435 &
  149.26 &
  High &
  \begin{tabular}[c]{@{}l@{}}Direct Sunlight (Morning)\\ Partly Cloudy (Noon)\\ Overcast (Afternoon)\end{tabular} &
  Yes &
  Yes \\ \cline{2-8} 
 &
  \textbf{Med School} &
  20700 &
  119.25 &
  Medium &
  \begin{tabular}[c]{@{}l@{}}Direct Sunlight (Morning)\\ Direct Sunlight (Noon)\\ Overcast (Afternoon)\end{tabular} &
  Yes &
  No \\
   \bottomrule
  
\end{tabular}%
}
\end{table*}

\begin{figure*}[!ht]
    \centering
    \includegraphics[width=0.75\linewidth]{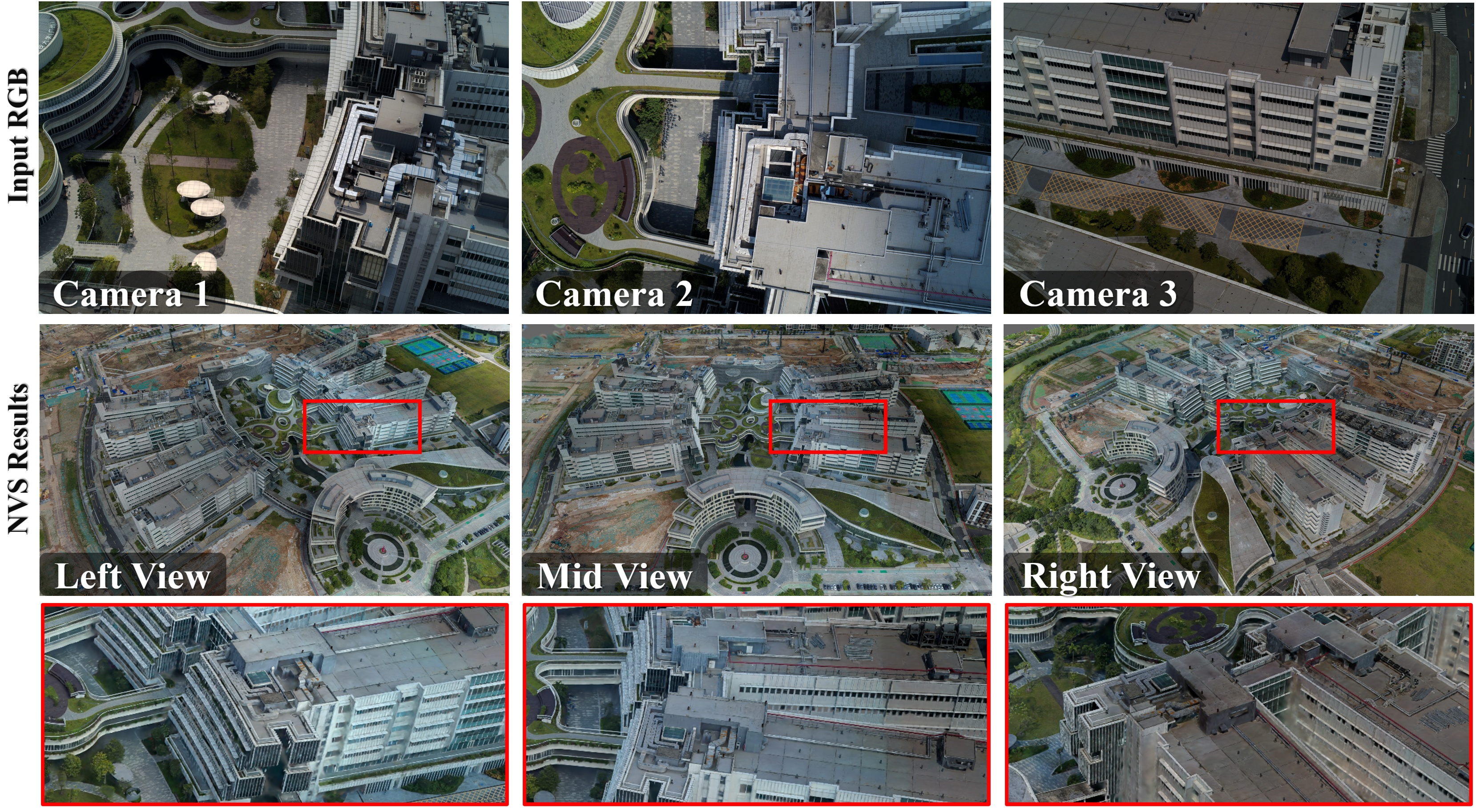}
    \caption{\textbf{Reconstruction challenges under partly cloudy illumination.}
In partly cloudy scenes, moving clouds create strongly varying local illumination, so the same region appears sunlit in some views and shadowed in others.
This breaks photometric consistency across viewpoints and leads to color inconsistencies.}

    \label{fig:illu}
\end{figure*}

\paragraph{Per-scene statistics.}
Tab.~\ref{tab:data_statsic} summarizes the per-scene statistics of our dataset. 
We cover ten urban regions grouped into three scales: compact campus-level blocks (\emph{Small Scale}), mid-sized institutional and residential districts (\emph{Medium Scale}), and city-block-level neighborhoods (\emph{Large Scale}). For each scene we report the total number of multi-view images aggregated over the three time slots, the approximate flight height, a categorical building-density label (\emph{Low/Medium/High}), the illumination configuration, and two binary attributes indicating the presence of large water bodies (\emph{Water Area}) and extensive glass façades (\emph{Glass Facade}).

Across all scenes, the total number of images per scene ranges from roughly 5k to over 20k, with nominal flight heights concentrated around 110-150~m to keep the ground sampling distance nearly consistent. Examples are illustrated in Fig.~\ref{fig:scenestat}. The building-density labels span two low-density scenes, five medium-density scenes, and three high-density scenes, covering open campuses, mixed-use blocks, and dense residential areas. The illumination types combine three time-of-day slots (morning, noon, afternoon) with varying sky conditions (direct sunlight, partly cloudy (see Fig.~\ref{fig:illu}), overcast), yielding sequences that range from purely clear-sky (e.g., \textit{Tec School}) to mixed clear/overcast setups (e.g., \textit{Gym}, \textit{Buildings}, \textit{Town}). The \emph{Water Area} and \emph{Glass Facade} columns highlight challenging scenes where strong reflections or transparency are prominent: six scenes contain sizable rivers or lakes, three scenes exhibit large glass façades, and two scenes (\textit{iPark} and \textit{Town}) feature both. In addition, the \textit{Med School} scene in particular contains a large-scale water body combined with complex surrounding structures, making it one of the most challenging cases in our benchmark. These attributes allow users to define targeted subsets focusing on specific difficulties such as specular water, glass buildings, complex architecture, or densely built environments.

\begin{table}[t]
    \centering
    \caption{Summary of Mean SfM alignment statistics and camera position uncertainty of 10 Scenes.}
    \label{tab:sfm_stats}
    \begin{tabular}{ll}
        \toprule
        \multicolumn{2}{c}{SfM alignment report} \\
        \midrule
        Average Projections & 13\,921\,072 \\
        Average track length & 3.3 \\
        Maximum non-compressible error [px] & 2.00 \\
        Median reprojection error [px] & 0.70 \\
        Mean reprojection error [px] & 0.79 \\
        \bottomrule
    \end{tabular}

    \vspace{0.8em}

    \begin{tabular}{lccc}
        \toprule
        \multicolumn{4}{c}{Relative camera position uncertainty [m]} \\
        \midrule
        & X & Y & Z \\
        \midrule
        Mean & $\leq$0.001 &	$\leq$0.002 &	$\leq$0.001 \\
        Standard deviation & $\leq$0.001 & $\leq$0.001 & $\leq$0.001 \\
        Maximum & $\leq$0.008 &	$\leq$0.019	& $\leq$0.014 \\
        Minimum & $\leq$0.001 & $\leq$0.001 & $\leq$0.001 \\
        \bottomrule
    \end{tabular}
\end{table}

\begin{figure}[!h]
    \centering
    \includegraphics[width=1\linewidth]{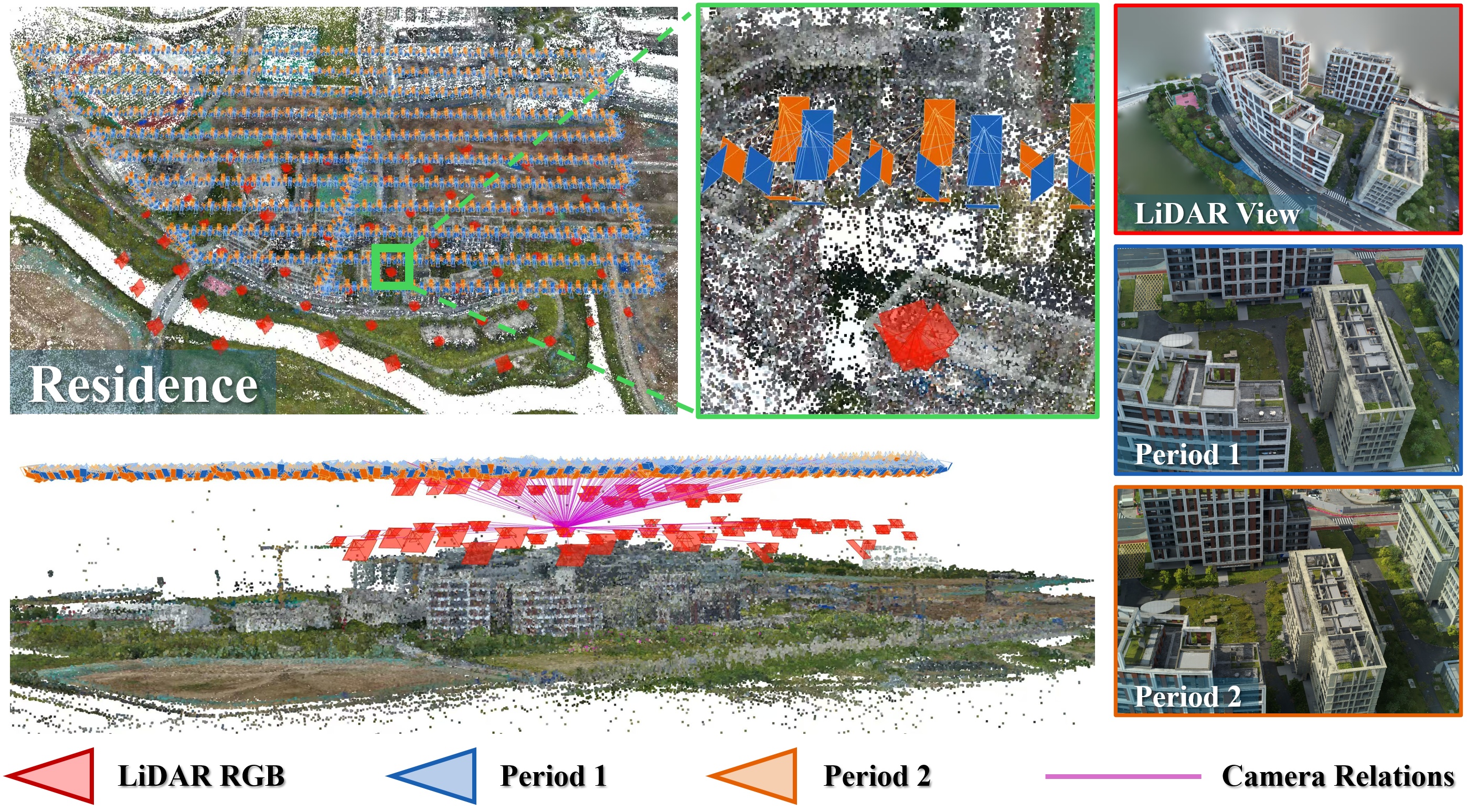}
    \caption{\textbf{LiDAR-Guided SfM Pipeline.}}
    \label{fig:sfm}
\end{figure}

\paragraph{Alignment statistics.} Fig.~\ref{fig:sfm} demonstrates an  example of the LiDAR-as-anchor for the SfM alignment for two time slots on \textit{Residence} Scene. 
And Table~\ref{tab:sfm_stats} summarizes the global SfM alignment quality and camera position uncertainty. The reconstruction is strongly constrained, with roughly $1.4\times10^{7}$ feature projections and an average track length of 3.3 observations per 3D point. The median and mean reprojection errors are 0.70~px and 0.79~px, both well below one pixel, indicating accurate camera calibration and bundle adjustment. The relative camera position uncertainty is on the order of millimeters in all three axes on average, with maximum uncertainties below 2~cm, demonstrating a geometrically stable camera network and a reliable  multi-temporal registration.
\section{Implementation Details}
\subsection{Parameter Settings}
For all 3DGS-based baselines evaluated under our single-GPU setting, we adopt a unified training schedule for fairness and stable reconstruction. Unless otherwise specified, each model is trained for 90{,}000 iterations in total. The densification interval is set to 300 iterations, the opacity reset interval is set to 9{,}000 iterations, and densification is disabled after 60{,}000 iterations. This allows the scene to be sufficiently populated and refined in the early and mid stages of training, while preventing unbounded growth of Gaussians and keeping memory usage under control on a single RTX A800 80~GB GPU.
\subsection{Training Strategy on Single GPU}
All methods in our benchmark are trained on a single RTX A800 80~GB GPU. 
Directly using the original training schedules leads to unfair behavior on large-scale scenes: 
standard 3DGS-style pipelines will reduce opacity many times and then prune low-opacity Gaussians at the next densification step (100 iteration). 
On a single GPU, however, a full sweep over all training views takes many iterations. 
As a consequence, many Gaussians are pruned before they have been sufficiently updated by all views, which results in over-pruning and loss of fine structures, especially in large or sparsely observed regions.

\begin{figure}[!h]
    \centering
    \includegraphics[width=1\linewidth]{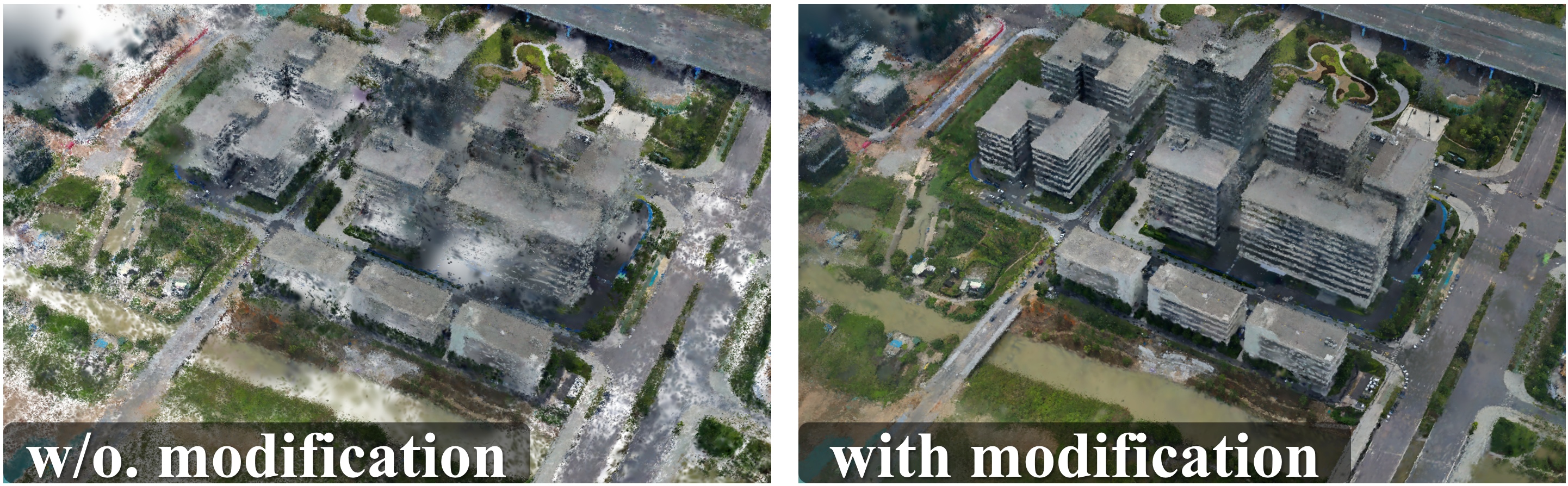}
    \caption{\textbf{Effect of the modified pruning schedule.}
Qualitative comparison  after the first opacity reset iteration (9000) and pruning. left: Opacity reset iteration is 9,000, densify interval is 300, and we visualized the result of 9,301 (9000+300+1) iteration; Right: Opacity reset iteration is 9,000, image count and densify interval is 1770, and we visualized the result of 10771 (9000+1770+1) iteration. }
    \label{fig:modify}
\end{figure}

To ensure both stability and fairness across methods under this resource constraint, we adopt the unified single-GPU training strategy summarized in Alg.~\ref{alg:optimization}. 
The key modifications are highlighted in \colorbox{OliveGreen!30}{green} in the pseudocode. 
Intuitively, this schedule enforces a simple but important constraint: 
every Gaussian must “see” all views at least once after its opacity is reset before it can be judged as low-opacity and removed. 
This prevents premature pruning caused by partial-view updates, reduces the risk of deleting valid geometry in large scenes, and yields more stable training trajectories across all compared methods on a single 80~GB GPU. 
As shown in the comparison Fig~\ref{fig:modify}, the modified schedule preserves thin structures and distant geometry significantly better than the original pruning behavior.

\begin{algorithm}[!h]
		\caption{Optimized Training Strategy on Single GPU\\
		$w$, $h$: width and height of the training images}
		\label{alg:optimization}
		\begin{algorithmic}
			\State $M \gets$ SfM Points	\Comment{Positions}
            \State \colorbox{OliveGreen!30}{$N \gets$ Images Count	}
			\State $i \gets 0$	\Comment{Iteration Count}
            \State  \colorbox{OliveGreen!30}{$Opacity Iter \gets 0$} \Comment{Record the Last Opacity Reset}
            \If{IsRefinementIteration($i$)}
            \If {\colorbox{OliveGreen!30}{$i - Opacity Iter > N$}}
			\ForAll{Gaussians $(\mu, \Sigma, c, \alpha)$ $\textbf{in}$ $(M, S, C, A)$}
			\If{$\alpha < \epsilon$ or IsTooLarge($\mu, \Sigma)$}	\Comment{Pruning}
			\State RemoveGaussian()	
			\EndIf
			\If{$\nabla_p L > \tau_p$} \Comment{Densification}
			\If{$\|S\| > \tau_S$}	\Comment{Over-reconstruction}
			\State SplitGaussian($\mu, \Sigma, c, \alpha$)
			\Else								\Comment{Under-reconstruction}
			\State CloneGaussian($\mu, \Sigma, c, \alpha$)
            \EndIf
            \If{$i \div opacity~reset~interval == 0  $}
            \State ResetOpacity()
            \State \colorbox{OliveGreen!30}{$Opacity Iter \gets i$} \Comment{Record the Last Opacity Reset Iteration}
            \EndIf
			\EndIf	
			\EndFor		
			\EndIf
            \EndIf
			\State $i \gets i+1$	
		\end{algorithmic}
	\end{algorithm}

\section{Benchmark Metrics}

\noindent\textbf{Novel view synthesis (NVS).}
Given a set of test images $\{I_n\}_{n=1}^N$ and the corresponding rendered images $\{\hat{I}_n\}_{n=1}^N$, we evaluate NVS quality using PSNR, SSIM, and LPIPS.

The peak signal-to-noise ratio (PSNR) is defined as:
\begin{equation}
    \mathrm{PSNR}(I, \hat{I}) = 10 \log_{10} \left( \frac{\mathrm{MAX}^2}{\mathrm{MSE}(I, \hat{I})} \right).
\end{equation}

Structural similarity (SSIM) between $I$ and $\hat{I}$ is defined as
\begin{equation}
    \mathrm{SSIM}(I, \hat{I}) =
    \frac{(2 \mu_I \mu_{\hat{I}} + C_1)(2 \sigma_{I\hat{I}} + C_2)}
         {(\mu_I^2 + \mu_{\hat{I}}^2 + C_1)(\sigma_I^2 + \sigma_{\hat{I}}^2 + C_2)}.
\end{equation}

LPIPS is defined as:
\begin{equation}
    \mathrm{LPIPS}(I, \hat{I}) =
    \sum_{l} w_l \left\| \hat{\phi}_l(I) - \hat{\phi}_l(\hat{I}) \right\|_2^2.
\end{equation}

\vspace{0.5em}
\noindent\textbf{Geometry accuracy.}
We evaluate geometric accuracy by comparing a predicted point set $P$ (e.g., sampled from the reconstructed mesh) with a ground-truth point set $G$.
For a point $p \in P$, let
\begin{equation}
    d(p, G) = \min_{g \in G} \| p - g \|_2,
\end{equation}
and analogously $d(g, P)$ for $g \in G$.
Given a distance threshold $\tau$, we define precision and recall as
\begin{equation}
    \mathrm{Precision}(\tau) =
    \frac{\left| \{ p \in P \mid d(p, G) < \tau \} \right|}{|P|},
\end{equation}
\begin{equation}
    \mathrm{Recall}(\tau) =
    \frac{\left| \{ g \in G \mid d(g, P) < \tau \} \right|}{|G|}.
\end{equation}
The F1-score is the harmonic mean of precision and recall:
\begin{equation}
    \mathrm{F1}(\tau) =
    \frac{2 \cdot \mathrm{Precision}(\tau) \cdot \mathrm{Recall}(\tau)}
         {\mathrm{Precision}(\tau) + \mathrm{Recall}(\tau)}.
\end{equation}
In all experiments we use a fixed threshold $\tau=0.25m,~0.5m,~0.75m$ for fair comparison across methods and scenes.

\paragraph{TCC implementation details.}
For completeness, we detail how the four TCC components in Sec.~\ref{sec:exp} are computed.
For each fixed test viewpoint $v_k$ and $T=3$ time slots $t \in \{1,2,3\}$, we render albedo images $A_t^{(k)} \in [0,1]^{H \times W \times 3}$.
We first form the temporal mean albedo
\begin{equation}
    \bar{A}^{(k)}(x,c) = \frac{1}{T} \sum_{t=1}^{T} A_t^{(k)}(x,c),
\end{equation}
where $x \in \Omega$ indexes pixels and $c \in \{1,2,3\}$ is the color channel.

We compute MAE and RMSE between each slot and the temporal mean and averaging over the three slots gives
\begin{equation}\small
    \overline{\mathrm{MAE}}^{(k)} =
    \frac{1}{T} \sum_{t=1}^{T} \mathrm{MAE}_t^{(k)},~~
    \overline{\mathrm{RMSE}}^{(k)} =
    \frac{1}{T} \sum_{t=1}^{T} \mathrm{RMSE}_t^{(k)}.
\end{equation}
We map these errors to $[0,1]$ consistency scores via
\begin{equation}
    \mathrm{TCC}^{(k)}_{\mathrm{MAE}} =
    1 - \mathrm{clip}_{[0,1]}\!\big(10\,\overline{\mathrm{MAE}}^{(k)}\big),
\end{equation}
\begin{equation}
    \mathrm{TCC}^{(k)}_{\mathrm{RMSE}} =
    1 - \mathrm{clip}_{[0,1]}\!\big(10\,\overline{\mathrm{RMSE}}^{(k)}\big),
\end{equation}
where $\mathrm{clip}_{[0,1]}(x)=\min(\max(x,0),1)$, ensures that extremely large errors are saturated and the resulting scores remain in the valid range $[0,1]$ without becoming negative.

Similarly, we compute SSIM and LPIPS between each slot and the temporal mean:
\begin{equation}\small
    s_t^{(k)} = \mathrm{SSIM}\big(A_t^{(k)}, \bar{A}^{(k)}\big), \quad
    \ell_t^{(k)} = \mathrm{LPIPS}\big(A_t^{(k)}, \bar{A}^{(k)}\big),
\end{equation}
and average over time,
\begin{equation}\small
    \overline{s}^{(k)} = \frac{1}{T} \sum_{t=1}^{T} s_t^{(k)}, \quad
    \overline{\ell}^{(k)} = \frac{1}{T} \sum_{t=1}^{T} \ell_t^{(k)}.
\end{equation}
The corresponding consistency scores are
\begin{equation}\small
    \mathrm{TCC}^{(k)}_{\mathrm{SSIM}} = \overline{s}^{(k)}, \quad
    \mathrm{TCC}^{(k)}_{\mathrm{LPIPS}} =
    1 - \mathrm{clip}_{[0,1]}\big(\overline{\ell}^{(k)}\big),
\end{equation}
since SSIM is already in $[0,1]$ while LPIPS is a distance that we invert.

These four components are then combined as defined in the main paper.

\section{Additional Results}

\begin{figure*}[htpb]
    \centering
    \includegraphics[width=0.95\linewidth]{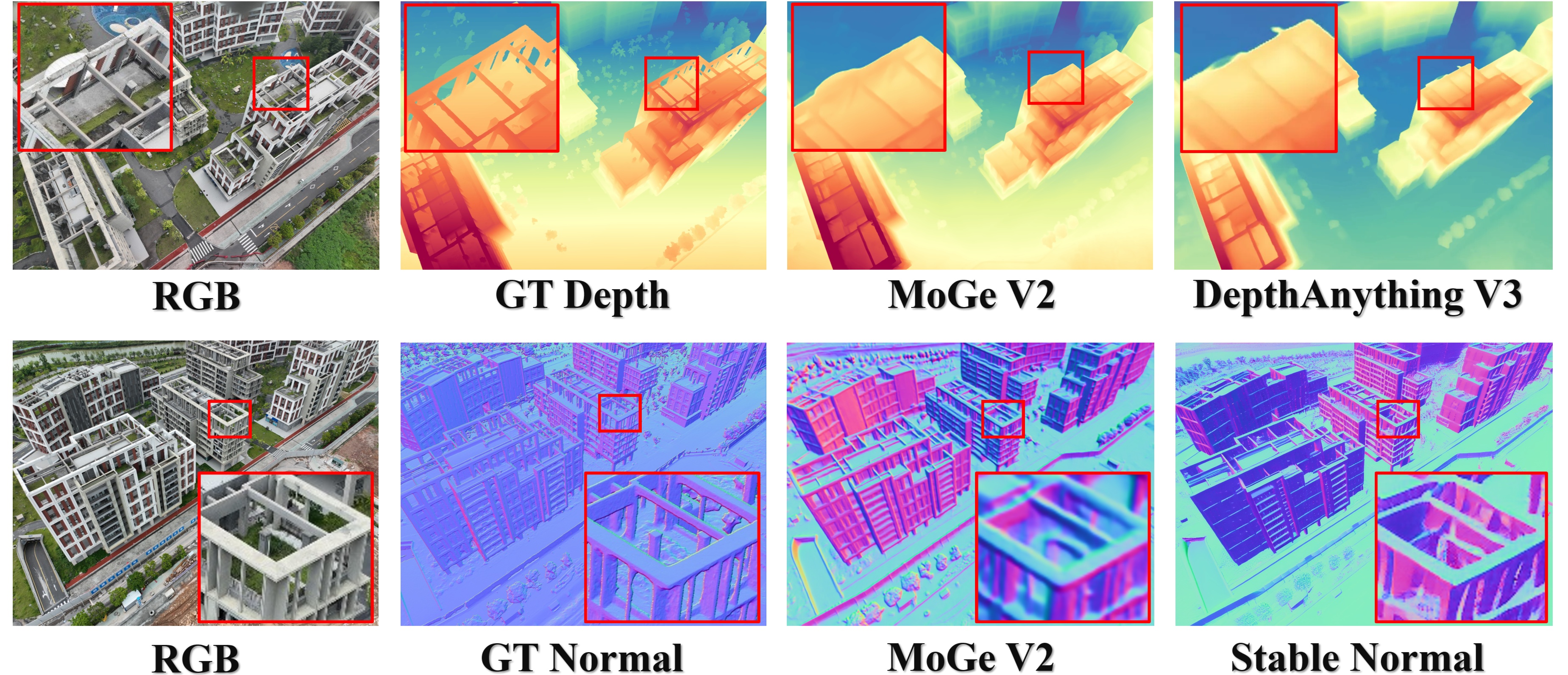}
    \vspace{-8pt}
    \caption{Qualitative comparison between SOTA monocular depth/normal estimators on SkyLume dataset.}
    \vspace{-8pt}
    \label{fig:mono_depth_normal}
\end{figure*}

We first compare our ground-truth depth and normals against state-of-the-art monocular depth \cite{wang2025moge2accuratemonoculargeometry, depthanything3} and  normal \cite{wang2025moge2accuratemonoculargeometry, ye2024stablenormal} estimators trained on generic indoor or street-view datasets. As illustrated in Fig.~\ref{fig:mono_depth_normal}, these methods struggle significantly under UAV oblique viewpoints: they systematically oversmooth façades and roof structures, blur depth discontinuities at building edges, and fail to recover thin elements such as railings, dormers, and small rooftop equipment. 

\begin{figure*}
    \centering
    \includegraphics[width=1\linewidth]{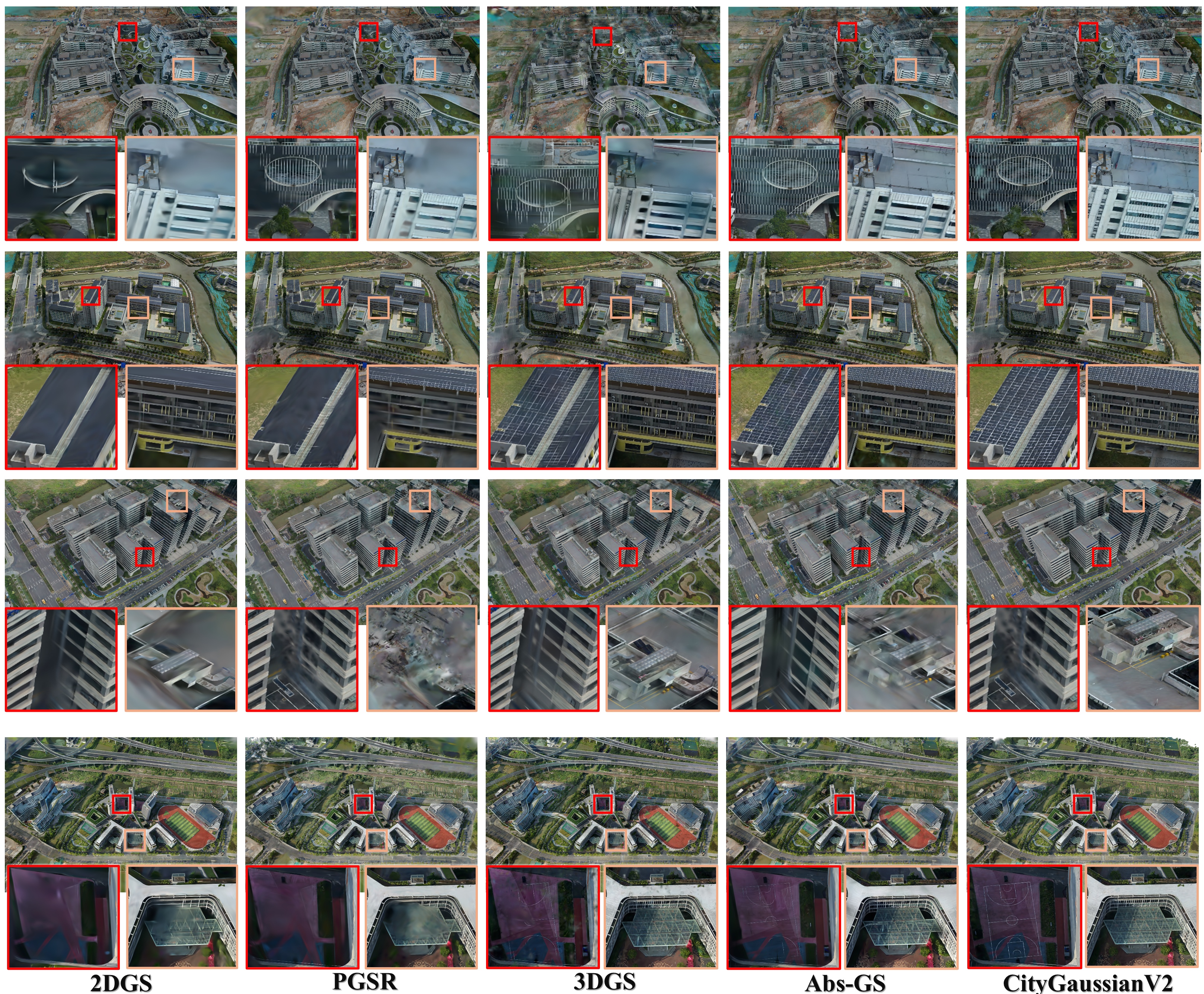}
    \caption{Additional bird's-eye visualizations rendered with the SuperSplat web viewer.}
    \label{fig:supersplat_birdview}
\end{figure*}

\begin{figure*}
    \centering
    \includegraphics[width=1\linewidth]{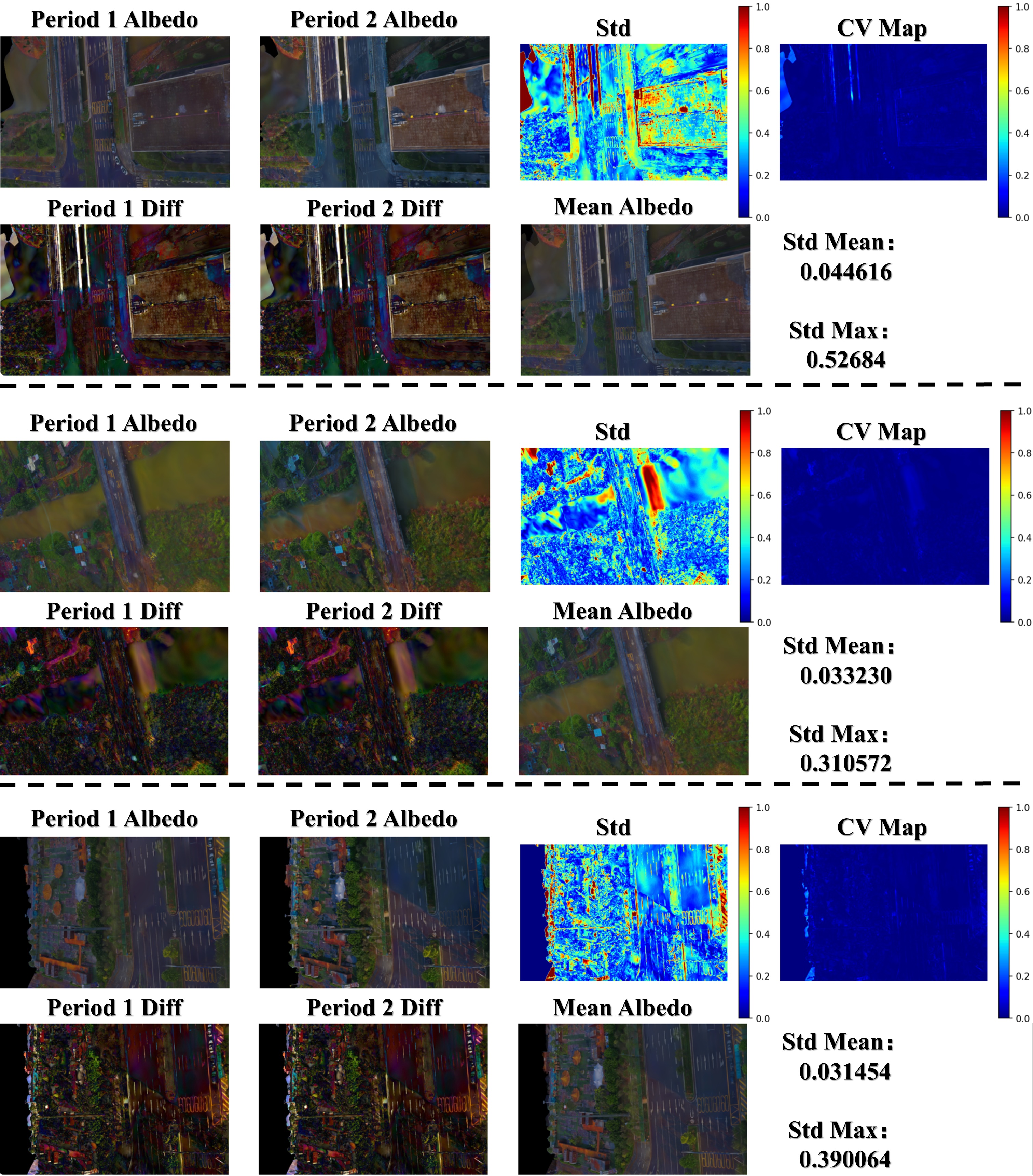}
    \caption{Additional TCC-albedo visualization of Ref-GS.}
    \label{fig:tcc_comparison_refgs}
\end{figure*}

\begin{figure*}
    \centering
    \includegraphics[width=1\linewidth]{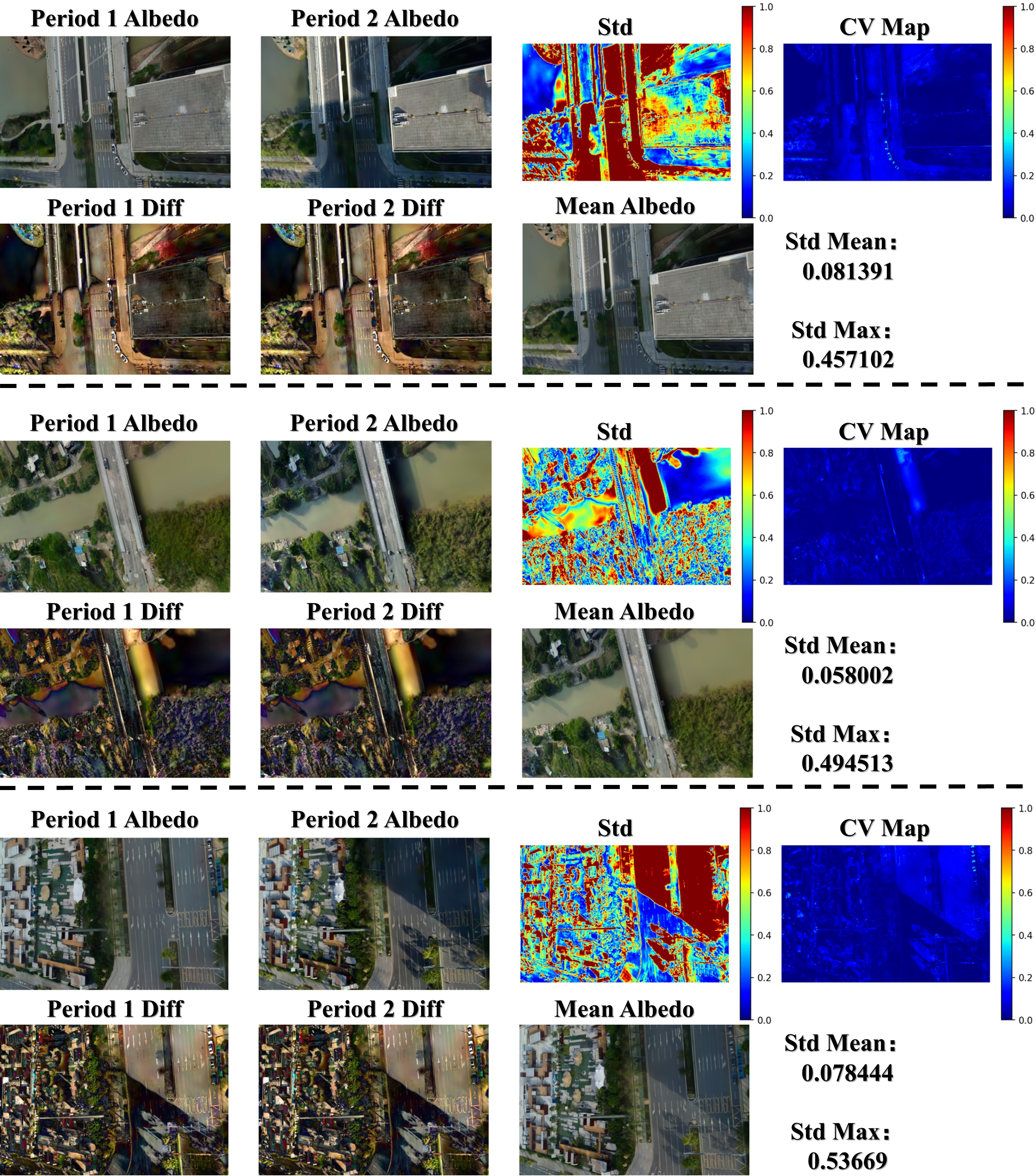}
    \caption{Additional TCC-albedo visualization of Ref-Gaussian.}
    \label{fig:tcc_comparison_refgaussian}
\end{figure*}

\begin{figure*}
    \centering
    \includegraphics[width=1\linewidth]{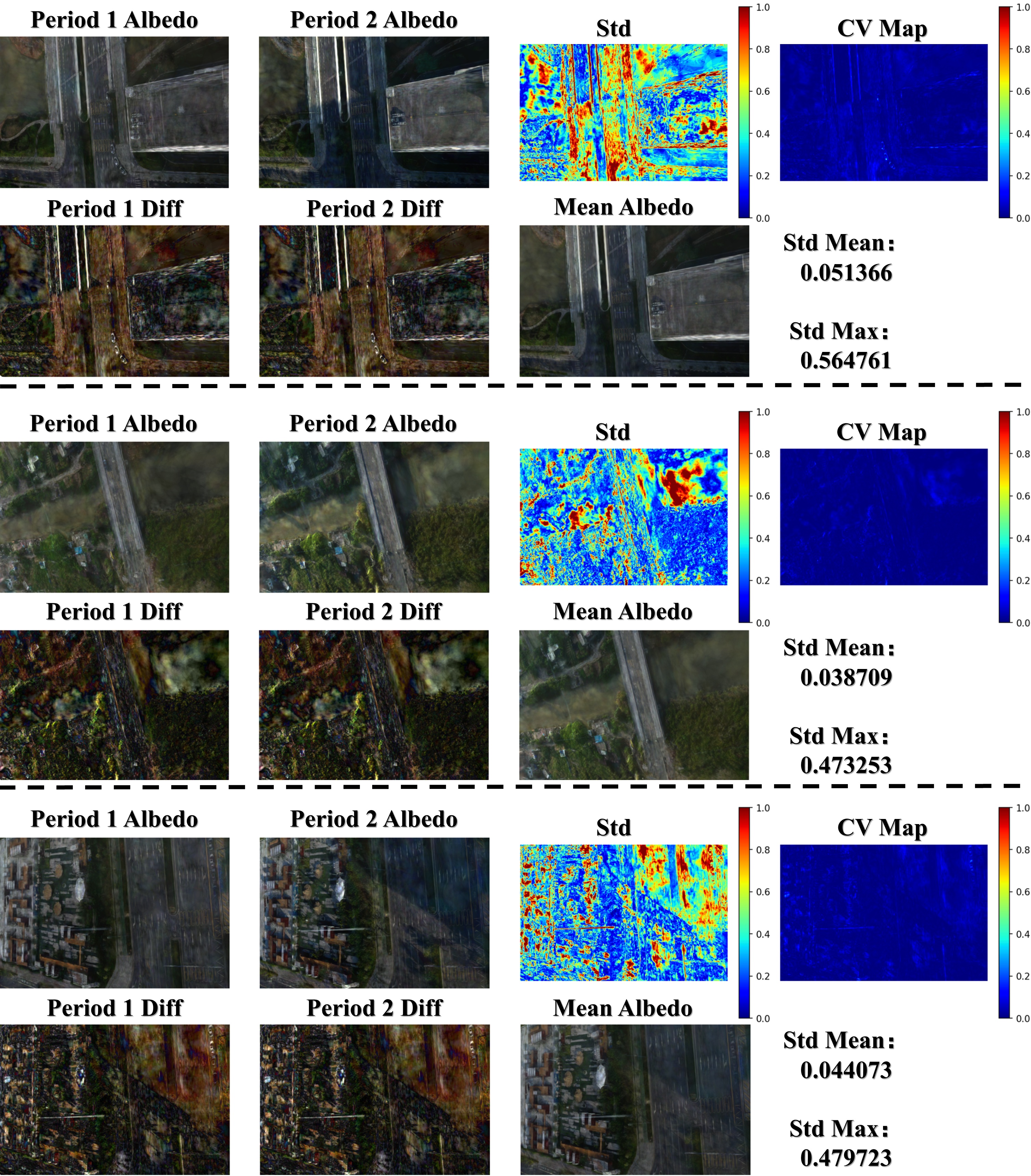}
    \caption{Additional TCC-albedo visualization of GS-IR.}
    \label{fig:tcc_comparison_gsir}
\end{figure*}

\begin{figure*}
    \centering
    \includegraphics[width=1\linewidth]{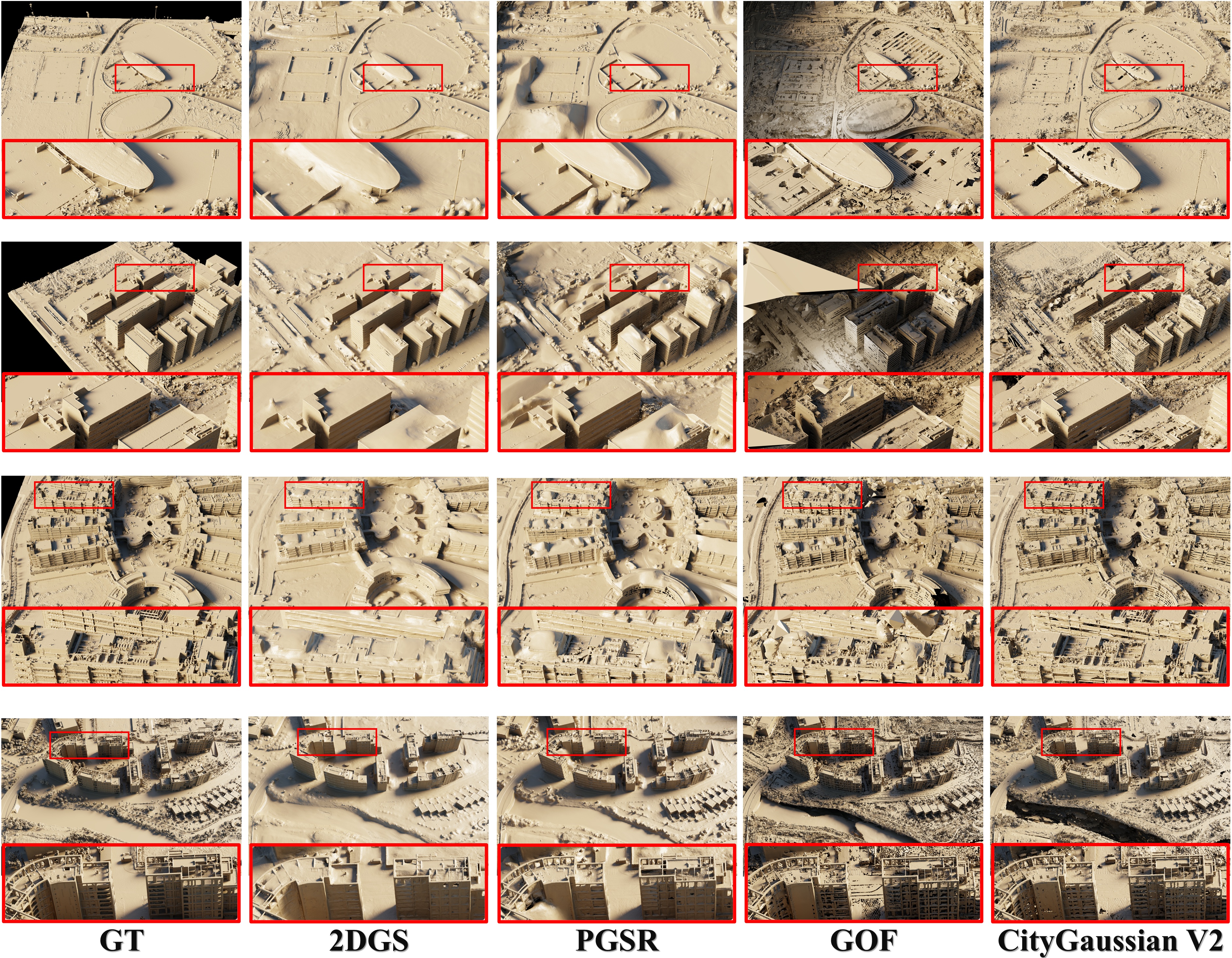}
    \vspace{-8pt}
    \caption{Additional bird's-eye geometry visualizations.}
    \vspace{-8pt}
    \label{fig:mesh2}
\end{figure*}

Fig.~\ref{fig:supersplat_birdview} shows bird's-eye visualizations rendered with the \textit{SuperSplat} web viewer for several large-scale scenes. We compare geometry-based methods 2DGS \cite{huang20242dgs} and PGSR \cite{chen2024pgsr} and observe that both methods exhibit under-reconstruction: distant buildings are partially missing, roof geometry is over-smoothed, and fine structures are either broken or entirely absent. For NVS-based baselines \cite{3dgs, ye2024absgs, liu2024citygaussianv2efficientgeometricallyaccurate}, strong illumination changes (e.g., moving cast shadows, partly cloudy conditions) lead to view-dependent artifacts such as ghosting along building contours and inconsistent shading across adjacent views. Quantitative and qualitative TCC-albedo of Ref-GS \cite{zhang2025ref-gs}, Ref-Gaussian \cite{yao2024ref-gaussian} and GS-IR \cite{liang2024gs-ir} are reported in Fig.~\ref{fig:tcc_comparison_refgs},~\ref{fig:tcc_comparison_refgaussian}, and~\ref{fig:tcc_comparison_gsir} highlighting the temporal instability of albedo estimates under challenging illumination, while geometry comparison are demonstrated in Fig.~\ref{fig:mesh2}.
\begin{figure*}
    \centering
    \includegraphics[width=1\linewidth]{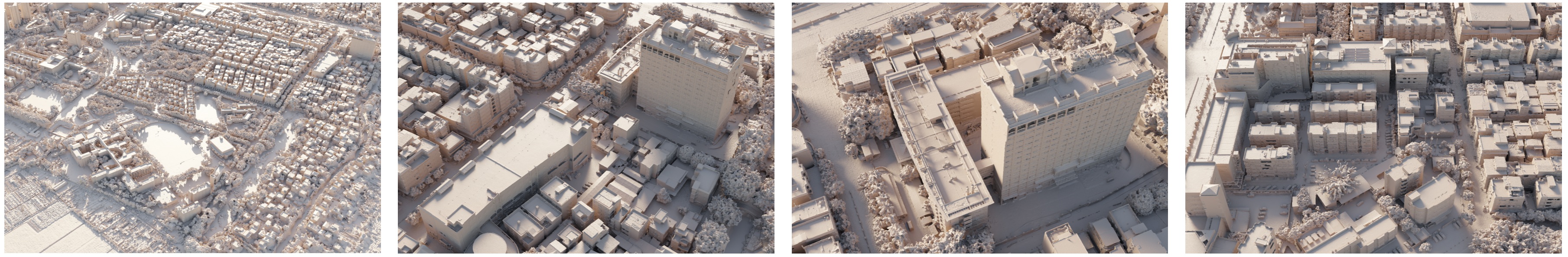}
    \caption{\textbf{Additional geometry ground-truth results.}
We visualize the geometry ground truth of the \textit{Town} scene, which features dense high-rise blocks, narrow streets, and deeply occluded courtyards. 
Our LiDAR-guided MVS pipeline produces a metrically accurate and topologically complete reference surface that preserves fine façade details and inner-structure layout, providing a reliable ground-truth benchmark for evaluating large-scale urban reconstruction methods in highly cluttered environments.}

    \label{fig:more_results}
\end{figure*}


\end{document}